\documentclass[journal]{IEEEtran}
\def\baselinestretch{1.2}

\usepackage{tikz}
\graphicspath{{images/}}
\usepackage{amsmath}
\usepackage{latexsym, amssymb}
\usepackage{graphicx}
\usepackage{amsmath, amsbsy}
\usepackage{amsopn, amstext}
\usepackage{ifpdf,hyperref}
\usepackage{cancel, color}
\usepackage{epstopdf}

\def\ttimes{\,\rotatebox[]{-90}{$\ltimes$}\,}

\def\tbtimes{\,\rotatebox[]{90}{$\bowtie$}\,}

\def\J{{\bf 1}}

\def\softmax{softmax}

\DeclareMathOperator{\Col}{Col}
\DeclareMathOperator{\Row}{Row}

\DeclareMathOperator{\lcm}{lcm}

\def\cal{\mathcal}

\def\ra{\rightarrow}

\def\a{\alpha}
\def\b{\beta}
\def\d{\delta}

\def\0{{\bf 0}}

\def\argmin{argmin}

\newcommand{\R}{{\mathbb R}}

\newcommand{\Z}{{\mathbb Z}}

\def\dsum{\mathop{\sum}\limits}

\newtheorem{thm}{Theorem}[section]
\newtheorem{dfn}[thm]{Definition}
\newtheorem{prp}[thm]{Proposition}
\newtheorem{exa}[thm]{Example}

\newtheorem{rem}[thm]{Remark}
\newtheorem{alg}[thm]{Algorithm}

\begin{document}

\title{On Dimension-Free Transformer:\\An Application of STP to AI}
\author{Daizhan Cheng
	\thanks{This work is supported partly by the National Natural Science Foundation of China (NSFC) under Grants 62073315.}
\thanks{Key Laboratory of Systems and Control, Academy of Mathematics and Systems Sciences, Chinese Academy of Sciences,
		Beijing 100190; and Research Center of Semi-tensor Product of Matrices P. R. China (e-mail: dcheng@iss.ac.cn).}
}


\maketitle

\begin{abstract} The matrix expressions for every parts of a transformer are firstly described. Based on semi-tensor product (STP) of matrices the hypervectors are reconsidered and the linear transformation over hypervectors is constructed by using projection. Its properties and calculating formulas are obtained. Using projection-based transformation of hypervector (PBTH), the framework  of dimension-free transformer (DFT) is proposed by verifying each linear transformation in a  transformer and replacing it by a proper PBTH, which allows the inputs and outputs being of arbitrary dimensions. Using balanced information about all entries, DFT must be more efficient in dealing  with signals.
\end{abstract}


\IEEEpeerreviewmaketitle

\section{Introduction}

In recent years the development of AI is unbelievably fast, and its impact on human lives is amazingly increasing. Particularly, the ChartGPT, the DeepSeek etc., as excellent examples of connectionism in AI, are shacking the AI community as well as the human society. The key engine in ChartGPT and DeepSeek etc. is the transformer, which is an attention-based neural network \cite{vas17}.

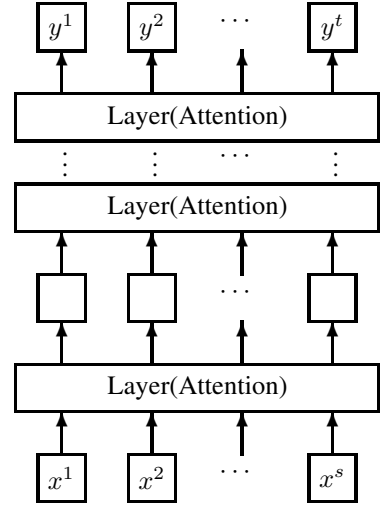
\begin{figure}
\centering
\setlength{\unitlength}{0.6 cm}
\begin{picture}(8,11)
\thicklines
\put(0.5,0){\framebox(1,1){$x^1$}}
\put(2.5,0){\framebox(1,1){$x^2$}}
\put(4.5,0.5){$\cdots$}
\put(6.5,0){\framebox(1,1){$x^s$}}
\put(0,2){\framebox(8,1){Layer(Attention)}}
\put(0.5,4){\framebox(1,1){}}
\put(2.5,4){\framebox(1,1){}}
\put(4.5,4.5){$\cdots$}
\put(6.5,4){\framebox(1,1){}}
\put(0,6){\framebox(8,1){Layer(Attention)}}
\put(0,8){\framebox(8,1){Layer(Attention)}}
\put(0.5,10){\framebox(1,1){$y^1$}}
\put(2.5,10){\framebox(1,1){$y^2$}}
\put(4.5,10.5){$\cdots$}
\put(6.5,10){\framebox(1,1){$y^t$}}
\put(1,1){\vector(0,1){1}}
\put(3,1){\vector(0,1){1}}
\put(5,1){\vector(0,1){1}}
\put(7,1){\vector(0,1){1}}
\put(1,3){\vector(0,1){1}}
\put(3,3){\vector(0,1){1}}
\put(5,3){\vector(0,1){1}}
\put(7,3){\vector(0,1){1}}
\put(1,5){\vector(0,1){1}}
\put(3,5){\vector(0,1){1}}
\put(5,5){\vector(0,1){1}}
\put(7,5){\vector(0,1){1}}
\put(1,9){\vector(0,1){1}}
\put(3,9){\vector(0,1){1}}
\put(5,9){\vector(0,1){1}}
\put(7,9){\vector(0,1){1}}
\put(1,7.2){$\vdots$}
\put(3,7.2){$\vdots$}
\put(7,7.2){$\vdots$}
\put(4.5,7.5){$\cdots$}
\end{picture}
\caption{Attenders in a Transformer \label{Fig.1.1}}
\end{figure}

Attentions are the key in a transformer.   Fig. \ref{Fig.1.1}, redrawing from \cite{wan24},  describes it.

From mathematical point of view, an attention is a multi-linear mapping
\begin{align}\label{1.1}
\pi: x^1\times \cdots \times x^s\ra y^1\times \cdots y^t.
\end{align}

\begin{itemize}
\item Case 1, Equal Length Sequences: $\dim(x_i)=\dim(y_j)$, $i\in [1,s]$, $j\in [1,t]$.  Then the mappings $\pi$ are commonly described by matrix multiplication (MatMul), or linear mapping (Linear) in component-wise.

\item Case 2,  Length-Varying Sequence: Say, $\dim{x_i}=m_i$, $i\in [1,s]$, $\dim(y_j)=n_j$, $j\in [1,t]$. In literature the most commonly used technique to deal with this case is using zero-padding. Than is, assume $m_0=\max_{1\leq i\leq s}m_i$ and $n_0=\max_{1\leq j\leq t}n_j$), then we pad $x^i$ by
$$
\tilde{x}^i=(x^i,\underbrace{0,\cdots,0}_{m_0-m_i})^{\mathrm{T}z},\quad i\in [1,s].
$$
Then the transformation defined in Case 1 can be used to deal with the padded date.  After the transformation, we received $\tilde{y}_j$ with length $n_0$. Then the so-called Padding-Mask is used to truncate it back to $y_j\in \R^{n_j}$.
\end{itemize}

The zero-padding approach has several weaknesses. (a) The additional zero-padded $0$ will cause some junk information into the resulting data. Then a padding mask is necessary to reduce the effects caused by such junk information.
(b) The components of a vector are not used in a balance way. There are some biases.
(c) The nominal $m_0$ and $n_0$ must be greater than or equal to the maximum $m_i$, $i\in [1,s]$ or maximum $n_j$, $j\in [1,t]$
respectively. Then the padded $\tilde{x}^i$, $i\in [1,s]$ (similarly,  $\tilde{y}^j$, $j\in [1,t]$) together become a sparse matrix, which also wast some memories.

This paper investigates the process on length-varying sequences. Instead of zero-padding approach, a projection-based padding is proposed, which can overcome the above mentioned weaknesses of zero-padding. Based on projection-based padding, a new transformer, called the dimension-free transformer (DFT) is proposed. In DFT, all kinds of linear transformations in a transformer have been reconsidered and replaced by corresponding  ``dimension-free" ones, which are essentially the projection-based padding and un-padding. In the sequel, you will see that the projection-based padding is also available for causal mask.

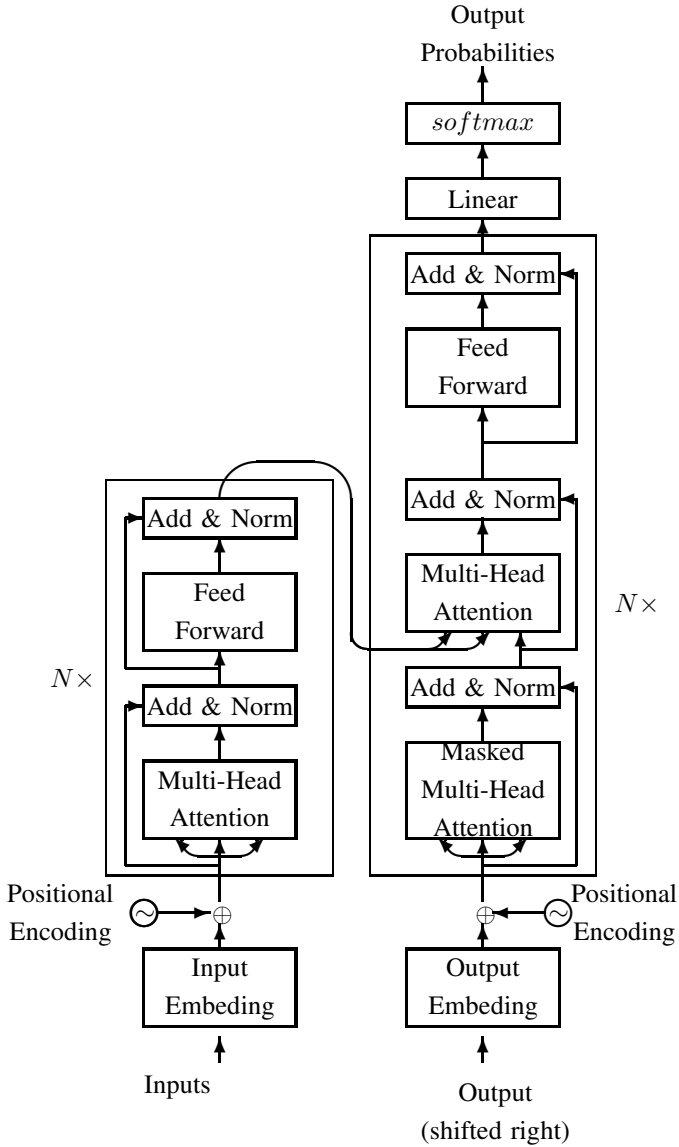
\begin{figure}
\centering
\setlength{\unitlength}{0.5 cm}
\begin{picture}(18,30)(-3,-1)
\thinlines
\put(0,6){\framebox(6,10.5){}}
\put(7,6){\framebox(6,17){}}
\thicklines
\put(1,2){\framebox(4,2){$\begin{array}{c}
\mbox{Input}\\ \mbox{Embeding}\end{array}$}}
\put(8,2){\framebox(4,2){$\begin{array}{c}
\mbox{Output}\\ \mbox{Embeding}\end{array}$}}
\put(1,7){\framebox(4,2){$\begin{array}{c}
\mbox{Multi-Head}\\ \mbox{Attention}\end{array}$}}
\put(8,7){\framebox(4,2.5){$\begin{array}{c}
\mbox{Masked}\\
\mbox{Multi-Head}\\ \mbox{Attention}\end{array}$}}
\put(1,10){\framebox(4,1){$
\mbox{Add \& Norm}$}}
\put(8,10.5){\framebox(4,1){$
\mbox{Add \& Norm}$}}
\put(1,12){\framebox(4,2){$\begin{array}{c}
\mbox{Feed}\\ \mbox{Forward}\end{array}$}}
\put(8,12.5){\framebox(4,2){$\begin{array}{c}
\mbox{Multi-Head}\\ \mbox{Attention}\end{array}$}}
\put(1,15){\framebox(4,1){$
\mbox{Add \& Norm}$}}
\put(8,15.5){\framebox(4,1){$
\mbox{Add \& Norm}$}}
\put(8,18.5){\framebox(4,2){$\begin{array}{c}
\mbox{Feed}\\ \mbox{Forward}\end{array}$}}
\put(8,21.5){\framebox(4,1){$
\mbox{Add \& Norm}$}}
\put(8,23.5){\framebox(4,1){$
\mbox{Linear}$}}
\put(8,25.5){\framebox(4,1){$
\softmax$}}
\put(8,28.2){$\begin{array}{c}
\mbox{Output}\\ \mbox{Probabilities}\end{array}$}
\put(2.8,4.75){$\oplus$}
\put(9.8,4.75){$\oplus$}
\put(1,5){\circle{0.7}}
\put(0.75,4.8){$\sim$}
\put(12,5){\circle{0.7}}
\put(11.75,4.8){$\sim$}
\put(11.65,5){\vector(-1,0){1.4}}
\put(1,5){\circle{0.7}}
\put(1.35,5){\vector(1,0){1.4}}
\put(3,5.3){\vector(0,1){1.7}}
\put(3,9){\vector(0,1){1}}
\put(3,14){\vector(0,1){1}}
\put(10,5.3){\vector(0,1){1.7}}
\put(3,1){\vector(0,1){0.7}}
\put(10,1){\vector(0,1){0.7}}
\put(3,4){\vector(0,1){0.7}}
\put(10,4){\vector(0,1){0.7}}
\put(1,0.2){$\mbox{Inputs}$}
\put(8,-0.5){$\begin{array}{c}
\mbox{Output}\\ \mbox{(shifted right)}\end{array}$}
\put(-3,4.8){$\begin{array}{c}
\mbox{Positional}\\ \mbox{Encoding}\end{array}$}
\put(-1.5,11){$N\times$}
\put(12,4.8){$\begin{array}{c}
\mbox{Positional}\\ \mbox{Encoding}\end{array}$}
\put(13.5,13){$N\times$}
\put(3,7){\oval(2,1)[b]}
\put(2.3,6.5){\vector(-1,1){0.5}}
\put(3.7,6.5){\vector(1,1){0.5}}
\put(10,7){\oval(2,1)[b]}
\put(9.3,6.5){\vector(-1,1){0.5}}
\put(10.7,6.5){\vector(1,1){0.5}}
\put(0.5,6.25){\line(1,0){2.5}}
\put(0.5,6.25){\line(0,1){4.25}}
\put(0.5,10.5){\vector(1,0){0.5}}
\put(12.5,6.25){\line(-1,0){2.5}}
\put(12.5,6.25){\line(0,1){4.75}}
\put(12.5,11){\vector(-1,0){0.5}}
\put(3,11){\vector(0,1){1}}
\put(0.5,11.5){\line(1,0){2.5}}
\put(0.5,11.5){\line(0,1){4}}
\put(0.5,15.5){\vector(1,0){0.5}}
\put(4.75,16){\oval(3.5,2)[t]}
\put(8.25,12.5){\oval(3.5,1)[b]}
\put(7.75,12.5){\oval(2.5,1)[b]}
\put(6.5,16){\line(0,-1){3.8}}
\put(8.72,12){\vector(1,1){0.5}}
\put(9.72,12){\vector(1,1){0.5}}
\put(11,11.5){\vector(0,1){1}}
\put(11,12){\line(1,0){1.5}}
\put(12.5,12){\line(0,1){4}}
\put(12.5,16){\vector(-1,0){0.5}}
\put(10,14.5){\vector(0,1){1}}
\put(10,16.5){\vector(0,1){2}}
\put(10,20.5){\vector(0,1){1}}
\put(10,22.5){\vector(0,1){1}}
\put(10,24.5){\vector(0,1){1}}
\put(10,26.5){\vector(0,1){1}}
\put(10,17.5){\line(1,0){2.5}}
\put(12.5,17.5){\line(0,1){4.5}}
\put(12.5,22){\vector(-1,0){0.5}}
\put(10,9.5){\vector(0,1){1}}
\end{picture}
\caption{ A Transformer\label{Fig.1.2}}
\end{figure}

Next, let us observe the whole structure of a transformer depicted by Fig. \ref{Fig.1.2}, which  is  redrawn from \cite{vas17}.

 There are only few nonlinear functions, which are:
$$
\begin{array}{l}
softmax,\quad (\mbox{probabilitization}),\\
Norm,\quad (\mbox{normalization}),\\
ReLU,\quad (\mbox{rectified linear Unit}), ~\mbox{etc.}
\end{array}
$$
All of them can be considered as ``self-adjustments" of argument vectors/matrices without any dimension change, where self-adjustment means no other data are necessary.  Hence there is no dimension dispute.

All the other operators can be considered as  compounded operators with
$$
\begin{array}{l}
MaxMul,\quad (\mbox{matrix product}),\\
Linear,\quad (\mbox{linear mapping}), \\
+,\quad (\mbox{addition, e.g., Feedforward}),\\
Mask,\quad (\mbox{padding-mask, casual-mask}),\\
PE,\quad (\mbox{positional encoding}).\\
\end{array}
$$

All these operators require certain dimension-matching conditions. The DFT is assumed to be a transformer with all linear operators, including MatMul, Linear, and $+$, are all dimension free.

\begin{figure}
\centering
\setlength{\unitlength}{0.5 cm}
\begin{picture}(12,6)
\thicklines
\put(1,1){\circle{1}}
\put(0.7,0.9){$x_n$}
\put(1,5){\circle{1}}
\put(0.7,4.9){$x_1$}
\put(11,1){\circle{1}}
\put(10.7,0.9){$y_m$}
\put(11,5){\circle{1}}
\put(10.7,4.9){$y_1$}
\put(5,0){\framebox(2,6){$\Sigma\;|\; f$}}
\put(7,1){\vector(1,0){3.5}}
\put(7,5){\vector(1,0){3.5}}
\put(1.5,1){\vector(1,0){3.5}}
\put(1.5,5){\vector(1,0){3.5}}
\put(1.5,1.5){\vector(1,1){3.5}}
\put(1.5,4.5){\vector(1,-1){3.5}}
\put(1,3){$\vdots$}
\put(11,3){$\vdots$}
\put(3,5.2){\begin{tiny}$w_{1,1}$\end{tiny}}
\put(3,0.5){\begin{tiny}$w_{n,m}$\end{tiny}}
\put(2,4.2){\begin{tiny}$w_{1,m}$\end{tiny}}
\put(2,1.6){\begin{tiny}$w_{n,1}$\end{tiny}}
\end{picture}
\caption{Single Layer Neural Network\label{Fig.0.2}}
\end{figure}
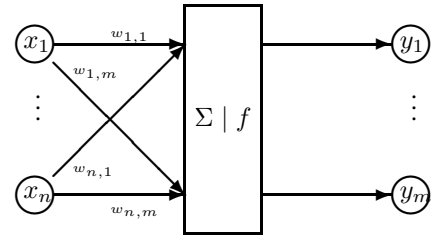

A single-layer neural network (NN) is described in Fig. \ref{Fig.0.2}. It can be formulated as

\begin{align}\label{1.2}
\begin{bmatrix}
y_1\\
\vdots\\
y_m
\end{bmatrix}
=f\left(W
\begin{bmatrix}
x_1\\
\vdots\\
x_n
\end{bmatrix}
+
\begin{bmatrix}
b_1\\
\vdots\\
b_m
\end{bmatrix}
\right),
\end{align}
where $W\in {\cal M}_{m\times n}$, $f(x)$: activation function, e.g., $f(x)=ReLU(x)=\max(0,x)$.

From the structure of single-layer NN one sees easily that the key part in an NN consists mainly of MatMul and matrix addition.
When the dimensions of matrices or vectors are varying, the DFT needs dimension-varying product and addition.

The semi-tensor product (STP) and semi-tensor addition (STA) are the generalizations of conventional matrix product and vector addition, which remove the dimension restriction of conventional matrices and vectors\cite{che12}. Hence STP and STA provide a proper tool to construct DFT.

Recently, using STP and STA, a dimension-free matrix theory has been developed\cite{che12}.
It has been used to many different fields. For instance, \cite{for16,muh16} introduced its application to Boolean networks; \cite{lu17} is a survey on its application to logical networks and finite valued systems; \cite{li18} surveyed its engineering applications; \cite{che21} surveyed its various applications to finite games; \cite{yan22} finite automata; and there are some works on cryptography \cite{zho18}, compressed sensing\cite{xie16}, etc.

The matrix-matrix (MM-) STP is a generalization of conventional matrix product, which allows the factor matrices of arbitrary dimensions; the matrix=vector (MV-) STP is a generalization of conventional matrix-vector product, which is also dimension-free; meanwhile, the STA, as a generalization of conventional vector addition, is an addition for vectors of different dimensions. These dimension-free operators from dimension-free matrix theory provide a set of suitable tools to construct DFT.

In a transformer, as well as in all AI based signal processes, it is commonly true that the data encountered are of higher order. Hence, describing them as hypermatrices is a proper way. In literature, the hypermatrices are also called the tensors. In this paper, we call them hypermatrices of order $k$. When the order of a set of data is $1$ we have a vector; when the order is $2$, we have a matrix; and as the order $k>2$ we have $k$-th order hypermatrices. We refer to \cite{lim13} for general hypermatrices, to \cite{cchen24} for some new applications of hypermatrices.

The paper is organized as follows. Section 2 reviews some basic concepts/results from dimension-free matrix theory, which are used in the sequel. They are (i) STP and STA; (ii) Cross-dimensional projection; (iii) Hypermatrices and Hypervectors.

(to be completed!)

Before ending this section, a list of notations is provided as follows.

\begin{itemize}

\item $\R$: set of real numbers.

\item $\Z_+$: set of positive integers.

\item $[a,b]$,  $a,b\in \Z_+$, $a\leq b$: set of integers $\{t\in \Z_+\;|\; a\leq t\leq b$.

\item ${\cal M}_{m\times n}$: set of $m\times n$ dimensional real matrices.

\item $\Col_i(A)$: $i$-th column of $A$.

\item $\Row_i(A)$: $i$-th row of $A$.

\item $I_{n}$: $n\times n$ Identity matrix.

\item $\d_n^i$: $\d_n^i=\Col_i(I_n)$.



\item ${\cal M}:=\bigcup_{m=1}^{\infty}\bigcup_{n=1}^{\infty}{\cal M}_{m\times n}$.

\item $\R^{\infty}:=\bigcup_{n=1}^{\infty}\R^{n}$.


\item $\R^{n_1\times \cdots\times n_d}$: Set of order $d$ and dimension $n_1\times \cdots\times n_d$ hypermatrices.

\item $\R^{n_1\ltimes \cdots\ltimes n_d}$: Set of order $d$ and dimension $n_1\times \cdots\times n_d$ hypervectors.


\item $\J_n$: ${\underbrace{[1,\cdots,1]}_n}^T$; $\J_{m\times n}$: $m\times n$ matrices with all entries being $1$.

\item$\otimes$: Kronecker product of matrices.


\item $\lcm(a,b)$: least common multiple of $a$ and $b$.
\item $\ltimes$: (standard) semi-tensor product.

\item $\ttimes$: dimension-keeping semi-tensor product.






%

\end{itemize}

\vskip 2mm

\section{Preliminaries}

\subsection{Semi-tensor Product and Semi-tensor Addition}

Denote the dimension-free Euclidian space as
$$
\R^{\infty}=\bigcup_{n=1}^{\infty}\R^n;
$$
and the dimension-free matrices as
$$
{\cal M}=\bigcup_{m=1}^{\infty}\bigcup_{n=1}^{\infty}{\cal M}_{m\times n}.
$$

\begin{dfn}\label{d2.1.1} Consider ${\cal M}$. A binary operator $\tbtimes: {\cal M}\times {\cal M}\ra {\cal M}$ is said to be a semi-tensor product (STP), if the followings are satisfied:
\begin{itemize}
\item[(1)] It is a generalization of conventional matrix product. That is, if $A\in {\cal M}_{*\times n}$ and $B\in {\cal M}_{n\times *}$, then
\begin{align}\label{2.1.1}
A\tbtimes B=AB.
\end{align}
\item[(2)] (Associativity)
\begin{align}\label{2.1.2}
(A\tbtimes B)\tbtimes C=A\tbtimes (B\tbtimes C).
\end{align}
\item[(3)] (Distributivity)   Let $A,B\in {\cal M}_{m\times n}$ and  $C\in {\cal M}$.
\begin{align}\label{2.1.3}
\begin{array}{l}
(A+B)\tbtimes C=(A\tbtimes C)+(B\tbtimes C);\\
C\tbtimes (A+B)=(C\tbtimes A)+(C\tbtimes B).
\end{array}
\end{align}
\end{itemize}
\end{dfn}

\begin{dfn}\label{d2.1.2}   Consider ${\cal M}$.  A binary operator $\oplus: {\cal D}\ra {\cal M}$ is said to be a semi-tensor addition (STA), where ${\cal D}\subset {\cal M}\times {\cal M}$, if the followings are satisfied:
\begin{itemize}
\item[(1)] (Commutativity)
\begin{align}\label{2.1.4}
A\oplus B=B\oplus A,\quad A,B\in {\cal D}.
\end{align}
\item[(2)] (Associativity)
\begin{align}\label{2.1.5}
(A\oplus B)\oplus C=A\oplus (B\oplus C),\quad A,B,C\in {\cal D}.
\end{align}
\end{itemize}
\end{dfn}

\begin{dfn}\label{d2.1.3}   Consider $\R^{\infty}$.  A binary operator $\oplus: \R^{\infty}\times \R^{\infty}\ra \R^{\infty}$ is said to be a semi-tensor addition (STA),  if the followings are satisfied:
\begin{itemize}
\item[(1)] (Commutativity)
\begin{align}\label{2.1.6}
x\oplus y=y\oplus x,\quad x,y\in \R^{\infty}.
\end{align}
\item[(2)] (Associativity)
\begin{align}\label{2.1.7}
(x\oplus y)\oplus z=x\oplus (y\oplus z),\quad x,y,z\in \R^{\infty}.
\end{align}
\end{itemize}
\end{dfn}

\begin{dfn}\label{d2.1.4}  Assume an STP $\tbtimes$ over ${\cal M}$ and an STA $\oplus$ over $\R^{\infty}$ are given.
A binary operator $\bowtie:{\cal M}\times \R^{\infty}\ra \R^{\infty}$ is called an STP (of matrix with vector), if the followings are satisfied:
\begin{itemize}
\item[(1)] (Semi-Group Property)
\begin{align}\label{2.1.8}
A\bowtie (B \bowtie x)=(A\bowtie B)\bowtie x,\quad A,B\in {\cal M},\; x\in \R^{\infty}.
\end{align}
\item[(2)] (Distributivity)
\begin{align}\label{2.1.9}
A\bowtie (x\oplus y)=(A\bowtie x)\oplus (A\bowtie y),\quad  A\in {\cal M},\; x,y\in \R^{\infty}.
\end{align}
\end{itemize}
\end{dfn}

Next, we introduce some commonly used STPs and STA.

\begin{dfn}\label{d2.1.5} The following are two commonly used STP of matrices.
\begin{itemize}
\item[(i)]  Let $A\in {\cal M}_{m\times n}$, $B\in {\cal M}_{p\times q}$, and $t=\lcm(n,p)$. A commonly used MM-STP of $A$ and $B$ is defined by \cite{che11}
\begin{align}\label{2.1.10}
A\ltimes B:=\left(A\otimes I_{t/n} \right)\left(B\otimes I_{t/p} \right) .
\end{align}
\item[(ii)] Let $A\in {\cal M}_{m\times n}$, $B\in {\cal M}_{p\times q}$, and $t=\lcm(n,p)$. The dimension-keeping (DK-) STP of $A$ and $B$ is defined by \cite{che24}
\begin{align}\label{2.1.11}
A\ttimes B:=\left(A\otimes \J^T_{t/n} \right)\left(B\otimes \J_{t/p} \right) .
\end{align}
\end{itemize}
\end{dfn}

\begin{dfn}\label{d2.1.5}
 Let $x\in \R^m$, $y\in \R^n$, and  $t=\lcm(m,n)$. The  STA of $x$ and $y$ is defined by \cite{che19}
\begin{align}\label{2.1.12}
x \pm y:=\left(x\otimes \J_{t/m}\right)\pm\left(y\otimes \J_{t/n} \right),
\end{align}
where we use $\pm$ to replace $\oplus$ ($\ominus$) respectively.
\end{dfn}

\begin{prp}\label{p2.1.6}  Let $A\in {\cal M}_{m\times n}$, $B\in {\cal M}_{p\times q}$, and $t=\lcm(n,p)$.  Consider the DK-STP defined by (\ref{2.1.11}). There exists a matrix
\begin{align}\label{2.1.13}
\Psi_{n\times p}:=\left(I\otimes \J^T_{t/n} \right)\left(I_m\otimes \J_{t/p} \right)\in {\cal M}_{n\times m},
\end{align}
called the bridge matrix, such that
\begin{align}\label{2.1.14}
M\ttimes N=M\Psi_{n\times p}N.
\end{align}
\end{prp}

\subsection{Cross-Dimensional Projection}

We refer to \cite{che19} for the projection over $\R^{\infty}$.

\begin{dfn}\label{d2.2.1} Let $x\in \R^m$, $y\in \R^n$, and $t=\lcm(m,n)$.
\begin{itemize}
\item[(i)] The inner product of $x$ and $y$ is
\begin{align}\label{2.2.1}
< x,y>_{{\cal V}}:=\frac{1}{t}<(x\otimes \J_{t/m}), (y\otimes \J_{t/n}>,
\end{align}
where $<\cdot,\cdot>$ is the conventional inner product on Euclidian space.
\item[(ii)] The norm  of $x\in \R^{\infty}$ is
\begin{align}\label{2.2.2}
\|x\|_{{\cal V}}:=\sqrt{< x,x>_{{\cal V}}}.
\end{align}
\item[(iii)] The distance of $x\in \R^m$ and $y\in \R^n$, is
\begin{align}\label{2.2.3}
d_{{\cal V}}(x,y):=\|x-y\|_{{\cal V}}.
\end{align}
\end{itemize}
\end{dfn}

\begin{dfn}\label{d2.2.2} Let $x\in \R^m$. The projection of $x$ to $\R^n$, denoted by $\pi^m_n(x)$, is defined as
\begin{align}\label{2.2.4}
\pi^m_n(x)=\argmin_{y\in \R^n}d_{{\cal V}}(x,y).
\end{align}
\end{dfn}

\begin{prp}\label{p2.2.3} $x\in \R^m$. The projection of $x$ to $\R^n$ is
\begin{align}\label{2.2.5}
\pi^m_n(x)=\Pi^m_nx,
\end{align}
where
\begin{align}\label{2.2.6}
\Pi^m_n=\frac{n}{t}\left(I_n\otimes \J^T_{t/n}\right)\left(I_m\otimes \J_{t/m}\right).
\end{align}
\end{prp}

Next, we define a new STA, called the nominal addition, on $\R^{\infty}$.

\begin{dfn}\label{d2.2.4} Let $x\in \R^m$, $y\in \R^n$. The $r$-nominal addition of $x$ and $y$ is defined by
\begin{align}\label{2.2.7}
x+_r y:=\Pi^m_r x+\Pi^n_r y\in \R^r.
\end{align}
\end{dfn}

We have the following result.

\begin{prp}\label{p2.2.5}

Assume  $t=\lcm(m,n)$, then
\begin{align}\label{2.2.8}
x+_r y:=\Pi^t_r (x+y)\in \R^r.
\end{align}
\end{prp}

\noindent{\it Proof.}
\begin{align}\label{2.2.9}
\begin{array}{l}
RHS_{(\ref{2.2.8})}=\Pi^t_r (x+y)\\
~=\Pi^t_r\left[(x\otimes \J_{t/m})+(y\otimes \J_{t/n})\right]\\
~=\Pi^t_r\left(x\otimes \J_{t/m}\right)+\Pi^t_r\left(y\otimes \J_{t/n}\right).\\
\end{array}
\end{align}

Hence, it is enough to show that
\begin{align}\label{2.2.10}
\Pi^t_r\left(x\otimes \J_{t/m}\right)=\Pi^m_r x;
\end{align}
and
\begin{align}\label{2.2.11}
\Pi^t_r\left(y\otimes \J_{t/n}\right)=\Pi^n_r y.
\end{align}

We prove (\ref{2.2.10}) first.

Assume $t=\lcm(m,n)=\mu m$, $s=\lcm(t,r)$,  $\d=\lcm(r,m)$, $s=\lambda \d$, $\xi=\lcm(r,n)$, and $s=\eta \xi$.
We calculate  $\Pi^t_r\left(x\otimes \J_{t/m}\right)$:
$$
\begin{array}{l}
\Pi^t_r\left(x\otimes \J_{t/m}\right)=\frac{1}{r}(I_r\otimes \J^T_{s/r})(I_t\otimes \J_{s/t})(x\otimes \J_{t/m})\\
~=\frac{r}{s} (I_r\otimes \J^T_{s/r}) \left[I_m\otimes (I_{\mu}\otimes \J_{s/t})\right](x\otimes \J_{t/m})\\
~=\frac{r}{s}(I_r\otimes \J^T_{s/r})\left[x\otimes (I_{\mu}\otimes \J_{s/t})(\J_{\mu})\right]\\
~=\frac{r}{s}(I_r\otimes \J^T_{s/r})\left[x\otimes (I_{\mu}\otimes \J_{s/t})(\J_{\mu}\otimes 1)\right]\\
~=\frac{r}{s}(I_r\otimes \J^T_{s/r})\left[x\otimes (\J_{\mu}\otimes \J_{s/t})\right]\\
~=\frac{r}{s}(I_r\otimes \J^T_{s/r})(x\otimes  \J_{s/m})\\
~=\frac{r}{s}(I_r\otimes \J^T_{s/r})(I_m\otimes  \J_{s/m})(x\otimes 1)\\
~=\frac{r}{s}(I_r\otimes \J^T_{\d/r}\otimes  \J^T_{\lambda})(I_m\otimes  \J_{\d/m}\otimes \J_{\lambda})x\\
~=\frac{r\lambda}{s}(I_r\otimes \J^T_{\d/r})(I_m\otimes  \J_{\d/m})x\\
~=\frac{r}{\d}(I_r\otimes \J^T_{\d/r})(I_m\otimes  \J_{\d/m})x\\
~=\Pi^m_r x.
\end{array}
$$
Similarly, we can prove (\ref{2.2.11}). The
(\ref{2.2.9}) follows.

\hfill $\Box$

\subsection{Hypermatrix and Hypervector}

\begin{dfn}\label{d2.3.1} \cite{lim13} Given $i_s\in [1,n_s]$, $s\in [1,d]$, a mapping $F:[1,n_1]\times\cdots\times[1,n_d]\ra \R$ is called a hypermatrix of order $d$ and dimension $n_1\times\cdots\times n_d$.

The set of hypermatrices of  order $d$ and dimension $n_1\times\cdots\times n_d$ is denoted by
$$
\R^{n_1\times n_2\times\cdots\times n_d}.
$$
\end{dfn}

Equivalently, A hypermatrix ${\cal H}\in  \R^{n_1\times n_2\times\cdots\times n_d}$ can be expressed as
\begin{align}\label{2.3.1}
{\cal H}=\{h_{i_1,i_2,\cdots,i_d}\;|\; i_s\in [1,n_s],~s\in [1,d]\}.
\end{align}

\begin{rem}\label{r2.3.2} It is clear from definition that a vector $x\in \R^n$ is a hypermatrix of order $1$, expressed  as
$$
{\cal X}=\{x_i\;|\; i\in [1,n]\}.
$$
A matrix $A\in {\cal M}_{m\times n}$ is a hypermatrix of order $2$, expressed as
$$
{\cal A}=\{a_{i,j}\;|\;i\in [1,m],~j\in[1,n]\}.
$$
\end{rem}

Next, we consider a matrix expression of a hypermatrix with order $d>2$. It is of fundamental importance in  analysis and manipulating hypermatrices.

An order $d$ hyper-matrix can be arranged into various different matrices as follows: Let ${\bf k}=\{k_1,k_2,\cdots,k_d\}$ be a set of $d$ indexes, which are used to label the elements of  a hyper-matrix ${\cal A}\in \R^{n_1\times \cdots\times n_d}$. ${\bf i}=\{i_1,i_2,\cdots,i_r\}$ and ${\bf j}=\{j_1,j_2,\cdots,j_s\}$, $r+s=d$ are two disjointed subset of  ${\bf k}$, such that,
$$
{\bf k}={\bf i}\bigcup {\bf j}
$$
becomes a partition. Then the matrix expression of ${\cal A}$ with respect to this partition is
\begin{align}\label{2.3.2}
A^{{\bf i}\times {\bf j}}\in {\cal M}_{n^{{\bf i}}\times n^{{\bf j}}},
\end{align}
where  $n^{{\bf i}}=\prod_{i\in {\bf i}}n_i$ and $n^{{\bf j}}=\prod_{j\in {\bf j}}n_j$. Precisely speaking, the elements of ${\cal A}$ are arranged into a matrix $A^{{\bf i}\times {\bf j}}$ in such a way: the rows of this matrix is labeled by  the index set ${\bf i}$ and its columns are labeled by the index set ${\bf j}$ in alphabetic order.

Note that each matrix expression uniquely determines  ${\cal A}$. Hence each matrix expression is equivalent to ${\cal A}$.
We refer to \cite{che23} for the conversion from one matrix form to another.

\begin{dfn}\label{d2.3.3} \cite{chepr} A vector $x\in \R^n$, $n=\prod_{i=1}^dn_i$ is called a hypervector of order $d$ and dimension $n_1\times \cdots \times n_d$, if there exist $x_i\in \R^{n_i}$, $i\in [1,d]$ such that
\begin{align}\label{2.3.3}
x=\ltimes_{i=1}^dx_i.
\end{align}

The set of hupervectors of order $d$ and dimension $n_1\times \cdots \times n_d$ is denoted by
$$
\R^{n_1\ltimes \cdots\ltimes n_d}.
$$
\end{dfn}

Note that the components of $x$ can then be expressed as
$$
\{x^1_{i_1}x^2_{i_2}\cdots x^d_{i_d}\;|\; i_s\in [1,n_s],~s\in [1,d]\},
$$
where $x^s_{i_s}$ is the $i_s$-th component of $x_s$. Then we may consider it as an element in $\R^{n_1\times \cdots\times n_d}$. Hence we have
\begin{align}\label{2.3.4}
\R^{n_1\ltimes \cdots\ltimes n_d}\subset \R^{n_1\times \cdots\times n_d}.
\end{align}

\section{Mathematical Formulation of a Transformer}

The model architecture of a  transformer is depicted by Fig. \ref{Fig.1.2}.

This section gives each part of transformer a rigorous mathematical description part-by-part.

\subsection{Input Embedding \& Positional Encoding}

First, we consider the Input Embedding.

Assume the input of a transformer is described in Fig. \ref{Fig.3.2}.

\begin{figure}
\centering
\setlength{\unitlength}{0.6 cm}
\begin{picture}(8,7)
\thicklines
\put(0,3){\line(1,0){8}}
\put(0,4){\line(1,0){8}}
\put(4,5.5){\vector(0,-1){1.4}}
\put(4,2.5){\vector(0,-1){1.4}}
\put(5.3,2.2){$s$}
\put(1.3,3.2){$x^1$}
\put(2.3,3.2){$\cdots$}
\put(3.3,3.2){$x^s$}
\put(4.3,3.2){$x^1$}
\put(5.3,3.2){$\cdots$}
\put(6.3,3.2){$x^s$}
\put(5.5,3){\oval(2.7,1)[b]}
\put(2.5,3){\oval(2.7,1)[b]}
\put(3.5,6){$NLP$}
\put(1,4.8){$Enbedding$}
\end{picture}
\caption{Input\label{Fig.3.2}}
\end{figure}

In Fig \ref{Fig.3.2},  after a NLP the signal is  converted into a sequence of vectors, denoted by $x_1,x_2,\cdots,x_s$, where $s$ is the batch size.

\begin{itemize}
\item Case 1, Fixed Dimension:
When $\dim(x^i)=n_0$, $i\in [1,s]$, The vectors can be arranged into a matrix row-by-row as
\begin{align}\label{3.1.1}
X=\begin{bmatrix}
x^1_1&x^1_2&\cdots&x^1_{n_0}\\
x^2_1&x^2_2&\cdots&x^2_{n_0}\\
\vdots&~&~&~\\
x^s_1&x^s_2&\cdots&x^s_{n_0}\\
\end{bmatrix}\in {\cal M}_{s\times n_0}.
\end{align}
We call (\ref{3.1.1}) the nominal form.

\item Case 2, Varying Dimension: Assume $n:=\max_{1\leq i\leq s}n_i\leq n_0$, where $n_0$ is preassigned, called the nominal dimension. Then we have
 \begin{align}\label{3.1.2}
X=\left[ \begin{array}{ccccccc}
x^1_1&x^1_2&\cdots&x^1_{n_1}&~&~&~\\
x^2_1&x^2_2&\cdots&~&x^2_{n_2}&~&~\\
\vdots&~&~&~\\
x^s_1&x^s_2&\cdots&0&0&x^s_{n_s}&~\\
\end{array}\right),
\end{align}
which is not a matrix, we call it a pseudo-matrix.

The existing method to deal with this situation is padding $X$ with zeros to make $X$ a matrix of dimension $s\times n_0$, called the zero-(token)-padding. Then the padded matrix becomes
 \begin{align}\label{3.1.3}
\tilde{X}=\begin{bmatrix}
x^1_1&x^1_2&\cdots&x^1_{n_1}&0&0&0\\
x^2_1&x^2_2&\cdots&~&x^2_{n_2}&0&0\\
\vdots&~&~&~\\
x^s_1&x^s_2&\cdots&0&0&x^s_{n_s}&0\\
\end{bmatrix}\in {\cal M}_{s\times n_0}.
\end{align}
\end{itemize}

\begin{rem}\label{r3.1.1} It is also natural to define
 \begin{align}\label{3.1.4}
X=
\begin{bmatrix}
x^1&x^2&\cdots&x^s\\
\end{bmatrix}.
\end{align}
But its transposed form (\ref{3.1.1}) or (\ref{3.1.2}) is commonly used in literature. We follow this convention.
\end{rem}

Next, we consider the generalization of queries ($Q$), keys ($K$), and values ($V$).

Hereafter in this section we consider only the nominal case. Under the nominal assumption we have $X,~Q,~K,~V\in {\cal M}_{s\times n_0}$ as
 \begin{align}\label{3.1.5}
\begin{array}{l}
Q^{\mathrm{T}}=W^qX^{\mathrm{T}},\\
K^{\mathrm{T}}=W^kX^{\mathrm{T}},\\
V^{\mathrm{T}}=W^vX^{\mathrm{T}},
\end{array}
\end{align}
where $W^q,~W^k, W^v\in {\cal M}_{n_0\times n_0}$.

Next, we consider the Positional Encoding. The fundamental formula is
\begin{align}\label{3.1.6}
\begin{array}{l}
PE(pos,2i)=\sin(pos/10000^{2i/d_{model}},\\
PE(pos,2i+1)=\cos(pos/10000^{2i/d_{model}}.\\
\end{array}
\end{align}

%
%

For notational ease, assume $d:=n_0=2r$. Then the position encoding is performed by
\begin{align}\label{3.1.7}
PE(X)=X+P,
\end{align}
where the position matrix $P$ is

\begin{align}\label{3.1.8}
\begin{array}{l}
P=\\
\left[
\begin{array}{ll}
\sin(0/10000^{0/d})&\cos(0/10000^{0/d})\\
\sin(1/10000^{0/d})&\cos(1/10000^{0/d})\\
\vdots &~\\
\sin((s-1)/10000^{0/d})&\cos((s-1)/10000^{0/d})\\
\end{array}\right.\\
\begin{array}{lll}
~\sin(0/10000^{1/d})&\cos(0/10000^{1/d})&\cdots\\
~\sin(1/10000^{1/d})&\cos(1/10000^{1/d})&\cdots\\
~\vdots&~&~\\
~\sin((s-1)/10000^{1/d})&\cos((s-1)/10000^{1/d})&\cdots\\
~\end{array}\\
\left.\begin{array}{ll}
\sin(0/10000^{r/d})&\cos(0/10000^{r/d})\\
\sin(1/10000^{r/d})&\cos(1/10000^{r/d})\\
~\vdots&~\\
\sin((s-1)/10000^{r/d})&\cos((s-1)/10000^{r/d})\\
\end{array}\right]\\
\end{array}
\end{align}

%
%
%
%
%
%

After positional encoding (PE), we have ds new vectors. Say, original $x^j$, $j\in [1,s]$  represent  words, and $p_k$, $k\in [1,d]$  represente  positions. Then after PE, we have compounded vectors
$$
x_j-p_k,\quad j\in [1,s], k\in [1,d],
$$
which represent word-positions. When the training data are large enough ($>ds$), the machine can learn all of them.

\begin{rem}\label{r3.1.2} The Output Embedding \& Positional Encoding is the same as the Input  Embedding \& Positional Encoding. Say in a translation learning, on the left hand side of the transformer (See Fig. \ref{Fig.3.2}),  the input might be the English sentences and for the  right hand side of the transformer, the input might be Chinese.
\end{rem}

\subsection{Scaled Dot-Product Attention}

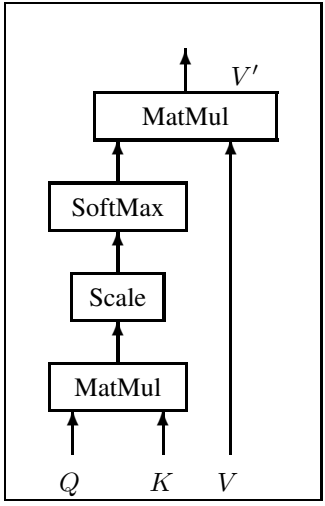
\begin{figure}
\centering
\setlength{\unitlength}{0.6 cm}
\begin{picture}(7,11)
\thinlines
\put(0,0){\framebox(7,11){}}
\thicklines
\put(1,2){\framebox(3,1){MatMul}}
\put(1.5,4){\framebox(2,1){Scale}}
\put(1,6){\framebox(3,1){SoftMax}}
\put(2,8){\framebox(4,1){MatMul}}
\put(1.5,1){\vector(0,1){1}}
\put(3.5,1){\vector(0,1){1}}
\put(5,1){\vector(0,1){7}}
\put(2.5,3){\vector(0,1){1}}
\put(2.5,5){\vector(0,1){1}}
\put(2.5,7){\vector(0,1){1}}
\put(4,9){\vector(0,1){1}}
\put(1.2,0.2){$Q$}
\put(3.2,0.2){$K$}
\put(4.7,0.2){$V$}
\put(5,9.2){$V'$}
\end{picture}
\caption{Scaled Dot-Product Attention\label{Fig.3.3}}
\end{figure}

As pointed by \cite{vas17}, in a transformer, the attention is ``all you need". That is, attentions are the key parts in a transformer. This subsection is devoted to build a dimension-free attention. A scaled dot-product attention is depicted as in Fig. \ref{Fig.3.3},  which  is also redrawn from \cite{vas17}.

In nominal case let $\dim(x_i)=n_0$, $i\in [1,s]$ This is the classical case, which has been discussed in literature so far. We briefly describe it as follows:

The input matrix $X$ is described as (\ref{3.1.1}), and the corresponding $Q$, $K$, and $V$ are obtained by (\ref{3.1.5}).

Then, the dot-product algorithm of the attention, depicted in Fig. \ref{Fig.3.3}, is expressed as
\begin{align}\label{3.2.1}
V'=Attension(Q,K,V)=AV,
\end{align}
where
\begin{align}\label{3.2.2}
A=\softmax\left(\frac{QK^T}{\sqrt{n}} \right).
\end{align}

\begin{figure}
\centering
\setlength{\unitlength}{0.5 cm}
\begin{picture}(16,8)
\thinlines
\put(0,0){\framebox(4,2){$V\in {\cal M}_{s\times n}$}}
\put(0,3){\framebox(4,2){$K\in {\cal M}_{s\times n}$}}
\put(0,6){\framebox(4,2){$Q\in {\cal M}_{s\times n}$}}
\put(10,4.5){\framebox(4,2){$A\in {\cal M}_{s\times s}$}}
\put(10,0){\framebox(4,2){$AV$}}
\put(14,1){\vector(1,0){2}}
\put(4,5.5){\oval(1,3)[r]}
\put(4.5,5.5){\vector(1,0){5.5}}
\put(4,1){\vector(4,1){7.8}}
\put(11.7,3){$\otimes$}
\put(12,4.5){\vector(0,-1){1}}
\put(12,2.7){\vector(0,-1){0.7}}
\put(5,6){$\softmax(\frac{QK^{\mathrm{T}}}{\sqrt{n}})$}
\put(15,1.2){$V'$}
\end{picture}
\caption{Dot-Product of Attention\label{Fig.3.4}}
\end{figure}
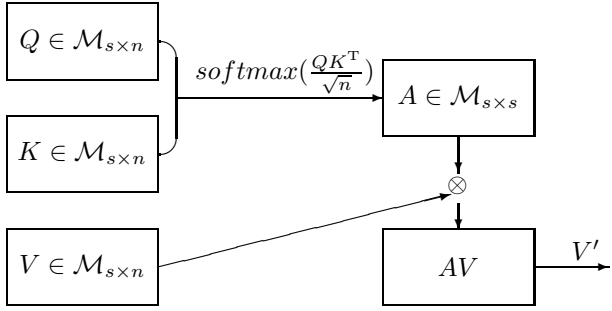

Fig. \ref{Fig.3.4} shows this.

\begin{rem}\label{r3.2.1} It is well known that \cite{goo16} if $x=(x_1,\cdots,x_k)\in \R^k$,
\begin{align}\label{3.2.3}
\softmax(x)=\frac{1}{\dsum_{i=1}^k e^{x_i}}[e^{x_i}, e^{x_2},\cdots,e^{x_k}]\in \R^k.
\end{align}
Now for a matrix $E:=\frac{QK^{\mathrm{T}}}{\sqrt{s}}$,
\begin{align}\label{3.2.4}
\softmax(E):=
\begin{bmatrix}
\softmax(\Row_1(E))\\
\softmax(\Row_2(E))\\
\vdots\\
\softmax(\Row_s(E))\\
\end{bmatrix}.
\end{align}
\end{rem}

\subsection{Multi-Head Attention}

\begin{figure}
\centering
\setlength{\unitlength}{0.6 cm}
\begin{picture}(11,13)(0,-2)
\thinlines
\put(1.4,1){\oval(2,0.8)[t]}
\put(1.8,1.4){\oval(2,0.8)[t]}
\put(2,1){\oval(0.8,1.2)[rb]}
\put(2.4,1.4){\oval(0.8,1.2)[rb]}
\put(5.4,1){\oval(2,0.8)[t]}
\put(5.8,1.4){\oval(2,0.8)[t]}
\put(6,1){\oval(0.8,1.2)[rb]}
\put(6.4,1.4){\oval(0.8,1.2)[rb]}
\put(9.4,1){\oval(2,0.8)[t]}
\put(9.8,1.4){\oval(2,0.8)[t]}
\put(10,1){\oval(0.8,1.2)[rb]}
\put(10.4,1.4){\oval(0.8,1.2)[rb]}
\put(5.4,5){\oval(10,0.8)[t]}
\put(5.8,5.4){\oval(10,0.8)[t]}
\put(10,4){\oval(0.8,1.2)[rb]}
\put(10.4,4.4){\oval(0.8,1.2)[rb]}
\put(10.4,5){\line(0,-1){1}}
\put(10.8,5.4){\line(0,-1){1}}
\put(1.4,1.4){\line(0,1){1.6}}
\put(1.8,1.8){\line(0,1){1.2}}
\put(5.4,1.4){\line(0,1){1.6}}
\put(5.8,1.8){\line(0,1){1.2}}
\put(9.4,1.4){\line(0,1){1.6}}
\put(9.8,1.8){\line(0,1){1.2}}
\put(5.4,5.4){\line(0,1){1.6}}
\put(5.8,5.8){\line(0,1){1.2}}
\thicklines
\put(1,0.6){\oval(2,0.8)[t]}
\put(1,0.4){\oval(2,0.8)[b]}
\put(0,0.4){\line(0,1){0.2}}
\put(2,0.4){\line(0,1){0.2}}
\put(5,0.6){\oval(2,0.8)[t]}
\put(5,0.4){\oval(2,0.8)[b]}
\put(4,0.4){\line(0,1){0.2}}
\put(6,0.4){\line(0,1){0.2}}
\put(9,0.6){\oval(2,0.8)[t]}
\put(9,0.4){\oval(2,0.8)[b]}
\put(8,0.4){\line(0,1){0.2}}
\put(10,0.4){\line(0,1){0.2}}
\put(5,4.6){\oval(10,0.8)[t]}
\put(5,3.4){\oval(10,0.8)[b]}
\put(0,3.4){\line(0,1){1.2}}
\put(10,3.4){\line(0,1){1.2}}
\put(3.5,7){\framebox(3,1){Concat}}
\put(3.5,9){\framebox(3,1){Linear}}
%
\put(1,3.8){Scaled Dot-Product Attention}
\put(0.2,0.2){Linear}
\put(4.2,0.2){Linear}
\put(8.2,0.2){Linear}
\put(1,1){\vector(0,1){2}}
\put(5,1){\vector(0,1){2}}
\put(9,1){\vector(0,1){2}}
\put(1,-1){\vector(0,1){1}}
\put(5,-1){\vector(0,1){1}}
\put(9,-1){\vector(0,1){1}}
\put(0.8,-1.8){$V$}
\put(4.8,-1.8){$K$}
\put(8.8,-1.8){$Q$}
\put(5,5){\vector(0,1){2}}
\put(5,8){\vector(0,1){1}}
\put(5,10){\vector(0,1){1}}
\end{picture}
\caption{Multi-Head Attention\label{Fig.3.5}}
\end{figure}
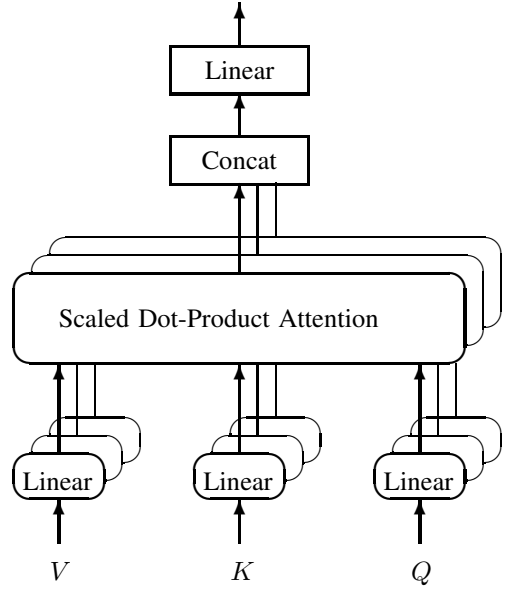


A multi-head attention is depicted in Fig. \ref{Fig.3.5}. Assume the head number is $n$. Then its function can be described in four steps.

\begin{itemize}
\item Step 1, Linear (Transformation): Through different linear transforms the inputs $V$, $K$, $Q$ can be converted into $\ell$ different forms.
\begin{align}\label{3.3.1}
\begin{array}{l}
V^i=T_v^i V,\\
K^i=T_k^i K,\\
Q^i=T_q^i Q,\quad i\in [1,\ell].
\end{array}
\end{align}

Here we do not need to require $\dim(v^i)$, $v^i\in V^i$ being of the same for different $i$.
These linear transformations are described in Fig. \ref{Fig.3.6}.

  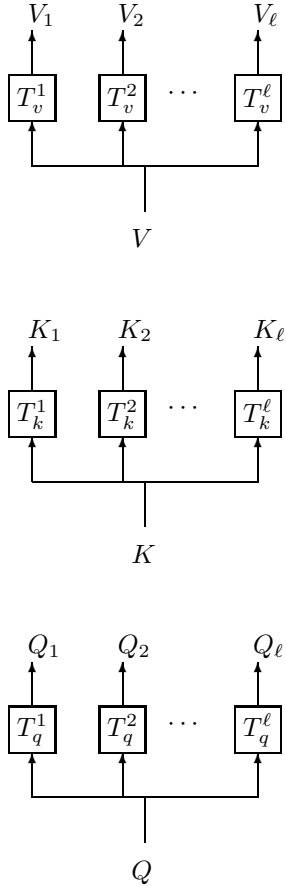
\begin{figure}
\centering
\setlength{\unitlength}{0.6 cm}
\begin{picture}(6,20)
\put(0,3){\framebox(1,1){$T^1_q$}}
\put(2,3){\framebox(1,1){$T^2_q$}}
\put(3.5,3.5){$\cdots$}
\put(5,3){\framebox(1,1){$T^{\ell}_q$}}
\put(0.5,2){\line(1,0){5}}
\put(0.5,2){\vector(0,1){1}}
\put(2.5,2){\vector(0,1){1}}
\put(5.5,2){\vector(0,1){1}}
\put(2.7,0.2){$Q$}
\put(3,1){\line(0,1){1}}
\put(0.5,4){\vector(0,1){1}}
\put(2.5,4){\vector(0,1){1}}
\put(5.5,4){\vector(0,1){1}}
\put(0.4,5.2){$Q_1$}
\put(2.4,5.2){$Q_2$}
\put(5.4,5.2){$Q_{\ell}$}
\put(0,10){\framebox(1,1){$T^1_k$}}
\put(2,10){\framebox(1,1){$T^2_k$}}
\put(3.5,10.5){$\cdots$}
\put(5,10){\framebox(1,1){$T^{\ell}_k$}}
\put(0.5,9){\line(1,0){5}}
\put(0.5,9){\vector(0,1){1}}
\put(2.5,9){\vector(0,1){1}}
\put(5.5,9){\vector(0,1){1}}
\put(2.7,7.2){$K$}
\put(3,8){\line(0,1){1}}
\put(0.5,11){\vector(0,1){1}}
\put(2.5,11){\vector(0,1){1}}
\put(5.5,11){\vector(0,1){1}}
\put(0.4,12.2){$K_1$}
\put(2.4,12.2){$K_2$}
\put(5.4,12.2){$K_{\ell}$}
\put(0,17){\framebox(1,1){$T^1_v$}}
\put(2,17){\framebox(1,1){$T^2_v$}}
\put(3.5,17.5){$\cdots$}
\put(5,17){\framebox(1,1){$T^{\ell}_v$}}
\put(0.5,16){\line(1,0){5}}
\put(0.5,16){\vector(0,1){1}}
\put(2.5,16){\vector(0,1){1}}
\put(5.5,16){\vector(0,1){1}}
\put(2.7,14.2){$V$}
\put(3,15){\line(0,1){1}}
\put(0.5,18){\vector(0,1){1}}
\put(2.5,18){\vector(0,1){1}}
\put(5.5,18){\vector(0,1){1}}
\put(0.4,19.2){$V_1$}
\put(2.4,19.2){$V_2$}
\put(5.4,19.2){$V_{\ell}$}
\end{picture}
\caption{Input Linear Mapping\label{Fig.3.6}}
\end{figure}

\item Step 2, Scaled Dot-Product Attention:   All heads do it separately. This process has been discussed before.
The outputs of this step are $V‘_i$, $i\in [1,\ell]$. We assume
\begin{align}\label{3.3.2}
\dim(v'_i)=r_i,\quad v'_i\in V'_i,\; i\in [1.\ell].
\end{align}

\item Step 3, Concat: For concatenation, we propose
\begin{align}\label{3.3.3}
\Col_j(V'):=\ltimes_{i=1}^{\ell} \Col_j(V'_i)\in \R^r,\quad j\in [1,s],
\end{align}
where $s$ is the batch size, and $r=\prod_{i=1}^{\ell}r_i$.

\item Step 4. Linear (Transformation):

Assume the output vectors are of dimension $r_0$ and the batch size is also $s$. Then we need a mapping
$$
W:\R^{r_1\ltimes \cdots \ltimes r_\ell\ltimes s}\ra \R^{r_0\ltimes s}.
$$
Hence, $W\in \R^{r_0\times s\times r_1\times \cdots \times r_{\ell}\times s}$ is a hypermatrix of the order $\ell+3$. It can be expressed as a matrix $M_W\in {\cal M}_{m\times n}$, where $m=r_0s$ and $n=rs$.
Then this linear transformation can be described as in Fig. \ref{Fig.3.7}

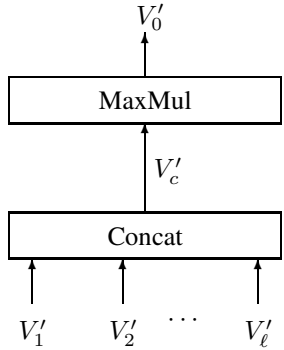
\begin{figure}
\centering
\setlength{\unitlength}{0.6 cm}
\begin{picture}(6,8)
\put(0,5){\framebox(6,1){MaxMul}}
\put(0,2){\framebox(6,1){Concat}}
\put(0.5,1){\vector(0,1){1}}
\put(2.5,1){\vector(0,1){1}}
\put(5.5,1){\vector(0,1){1}}
\put(3.5,0.5){$\cdots$}
\put(0.2,0.2){$V'_1$}
\put(2.2,0.2){$V'_2$}
\put(3.2,3.8){$V'_c$}
\put(5.2,0.2){$V'_{\ell}$}
\put(3,3){\vector(0,1){2}}
\put(3,6){\vector(0,1){1}}
\put(2.8,7.2){$V'_0$}

\end{picture}
\caption{Concat \& Linear Mapping\label{Fig.3.7}}
\end{figure}

\end{itemize}

\subsection{Mask}

There are two kinds of Masks in literature.

\begin{itemize}
\item[(i)] Padding mask: Padding mask is used for eliminating the affection of padded zeros. In DFT, since there is no padded zero, there is no demand for Padding mask.

\item[(ii)] Causal Mask (or Look-ahead Mask): Causal Mask is used to prevent the system to use future information. The Padding Matrix is
$$
PM=(p_{i,j})\in {\cal M}_{s\times s},
$$
where
\begin{align}\label{3.4.1}
p_{i.j}=
\begin{cases}
0,\quad j>i,\\
-\infty,\quad j\leq i.
\end{cases}
\end{align}

Then in classical case we have
\begin{align}\label{3.4.2}
Attention(Q,K,V)=softmax(\frac{QK^T}{\sqrt{s}}+PM)V.
\end{align}
%
%

\subsection{Add \& Norm}

There are two steps for Add \& Norm.

\begin{itemize}
\item Step 1, Add: Add is completed by residual network, which is depicted in Fig. \ref{Fig.3.8}.

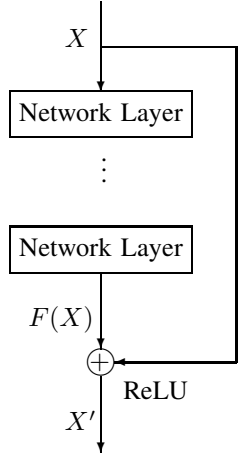
\begin{figure}
\centering
\setlength{\unitlength}{0.6 cm}
\begin{picture}(6,10)(0,-1)
\put(0,3){\framebox(4,1){Network Layer}}
\put(0,6){\framebox(4,1){Network Layer}}
\put(2,5){$\vdots$}
\put(2,9){\vector(0,-1){2}}
\put(2,3){\vector(0,-1){1.7}}
\put(2,1){\circle{0.6}}
\put(1.75,0.85){$+$}
\put(2,0.7){\vector(0,-1){1.7}}
\put(5,1){\vector(-1,0){2.7}}
\put(5,1){\line(0,1){7}}
\put(5,8){\line(-1,0){3}}
\put(2.5,0.2){ReLU}
\put(1.2,8){$X$}
\put(0.4,1.8){$F(X)$}
\put(1.2,-0.5){$X'$}
\end{picture}
\caption{Residual Add\label{Fig.3.8}}
\end{figure}

In nominal casee we assume
$$
X,~F(X)\in \R^{m_1\ltimes \cdots \ltimes m_s}.
$$
 Then the addition is defined as

\begin{align}\label{3.5.1}
X+F(X)=(x^i_j+f^i_j),
\end{align}
where $f^i_j$ is the  $j$-th element of $f^i$, which is the $i$-th component of $F(X)$.

Then we have
\begin{align}\label{3.5.2}
X'=ReLU(X+ F(X))
\end{align}

Note that ReLU is called the ``rectified linear unit", which is defined by
\begin{align}\label{3.5.3}
ReLU(x)=\max(0,x).
\end{align}

\item Step 2, Normalization: Normalization can be realized either vector-wise or layer-wise.
The formula is
\begin{align}\label{3.5.4}
y=\frac{x-E(x)}{\sqrt{Var(x)+\epsilon}}*\gamma + \beta,
\end{align}
where $0<\epsilon<<1$ is a small positive number (say, $1E-3$) to avoid zero denominator, $\gamma$ and $\beta$ are adjustable parameters.

Let
$$
X=(x_{i,j})\in {\cal M}_{d\times s},
$$
where $d=\dim(x)$, $x\in X$ and $s$ is the band size. When the vector-wise normalization is performed,
then for column $j$:

$x=\Col_j(X)$. $y=\Col_j(Y)$, $j\in [1,s]$, and
$$
\begin{array}{l}
E_j(x)=\frac{1}{d}\dsum_{i=1}^dx_{i,j}\\
Var(x)=\frac{1}{d} \sqrt{\dsum_{i=1}^d(x_{i,j}-E_j(x))^2}.
\end{array}
$$

 When the layer-wise normalization is performed:

$x=X$. $y=Y$,  and
$$
\begin{array}{l}
E(x)=\frac{1}{ds}\dsum_{i=1}^d\dsum_{j=1}^sx_{i,j}\\
Var(x)=\frac{1}{ds} \sqrt{\dsum_{i=1}^d\dsum_{j=1}^s(x_{i,j}-E(x))^2}.
\end{array}
$$

\end{itemize}

\subsection{Feed Forward}

The feed-forward network considers of two linear transformations (LinTra) with a ReLU activation in between, which is depicted
in Fig. \ref{Fig.3.9}.

\begin{figure}
\centering
\setlength{\unitlength}{0.6 cm}
\begin{picture}(4,9)
\put(0,2){\framebox(4,1){LinTra}}
\put(0,4){\framebox(4,1){ReLU}}
\put(0,6){\framebox(4,1){LinTra}}
\put(1.2,8.2){$FFN(X)$}
\put(1.8,0.2){$X$}
\put(2,1){\vector(0,1){1}}
\put(2,3){\vector(0,1){1}}
\put(2,5){\vector(0,1){1}}
\put(2,7){\vector(0,1){1}}
\end{picture}
\caption{Feed Forward\label{Fig.3.9}}
\end{figure}
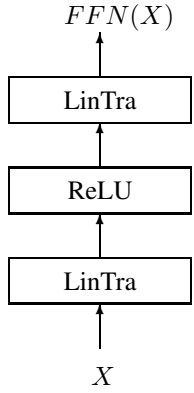

The Feed Forward is formulated as
\begin{align}\label{3.6.1}
FFN(X)=W_2 ReLU(W_1X+_{\a} B_1)+_{\b}B_2,
\end{align}
where $X\in {\cal M}_{s\times n_0}$, $W_1\in {\cal M}_{\a\times s}$, and $W_2\in {\cal M}_{\b\times \a}$.

\end{itemize}

\section{Generalized HyperVectors}

\subsection{Expressions of  Hypervectors}

To describe the objects through the  attention-based mappings, as well as some other operators in a transformer, we need a more general hypervectors, which are defined as follows.

\begin{dfn}\label{d4.1.1} A hypervector is an ordered finite set of finite dimensional vectors. Say,
\begin{align}\label{4.1.1}
X=\{x^1,\cdots,x^s\}, \quad x^i\in \R^{n_i},\;i\in [1,s].
\end{align}

The set of hypervectors $X$, satisfying (\ref{4.1.1}), is denoted by $\R^{n_1\ltimes \cdots\ltimes n_s}$.
\end{dfn}

Assume $X\in \R^{n_1\ltimes \cdots\ltimes n_s}$. It has three representation forms.

\begin{itemize}
\item[(i)] Pseudo-Matrix Form:
\begin{align}\label{4.1.2}
M_X:=\left[\begin{array}{cccccc}
x^1_1&x^1_2&\cdots&x^1_{n_1}&~&~\\
x^2_1&x^2_2&\cdots&\cdots&\cdots&x^1_{n_2}\\
\vdots&~&~&~&~&~\\
x^s_1&x^s_2&\cdots&\cdots&x^s_{n_s}&~\\
\end{array}
\right)
\end{align}
where the row number of $M_X$ is $s$ and the column numbers of $M_X$ are $n_i$, $i\in [1,s]$. That is, $\dim(\Row_i(X))=n_i$.

It is obvious that when $n_i=n_0$, $i\in [1,s]$, the pseudo-matrix form becomes a matrix. We call such $X$ homogeneous sequence.

\item[(ii)] Addition Form:
\begin{align}\label{4.1.3}
V_X=\begin{bmatrix}
x^1\\
x^2\\
\vdots\\
x^s
\end{bmatrix}\in \R^{n_a},
\end{align}
where $n_a=\dsum_{i=1}^s n_i$.

\item[(iii)] Product Form:
\begin{align}\label{4.1.4}
x=\ltimes_{i=1}^sx_i\in\R^{n_p},
\end{align}
where $n_p=\prod_{i=1}^s n_i$.
\end{itemize}

\begin{rem}\label{r4.1.1}
\begin{itemize}
\item[(i)] (\ref{4.1.4}) is the classical definition in Definition \ref{d2.3.3}.

\item[(ii)] It is easy to see that the expressions (\ref{4.1.1}), (\ref{4.1.2}), and (\ref{4.1.3}) are equivalent.
As for (\ref{4.1.4}), it is easy to see that   (\ref{4.1.1}) (or (\ref{4.1.2}) or  (\ref{4.1.3}) ) implies (\ref{4.1.4}). But from (\ref{4.1.4}) there might be many different hypervectors of the form of   (\ref{4.1.1}) (or (\ref{4.1.2}) or  (\ref{4.1.3}) ), which have the same product form as (\ref{4.1.4}).  The following proposition reveals their relationship. Moreover, we also refer to \cite{chepr2} for solving $x^i$, $i\in [1,s]$ from $x$.
\end{itemize}
\end{rem}

\begin{prp}\label{p4.1.2} \cite{chepr2}
Assume $0\neq x=\ltimes_{i=1}^sx_i=\ltimes_{i=1}^sy_i$, where $x_i,y_i\in \R^{n_i}$, $i\in [,s]$. Then
\begin{align}\label{4.1.5}
y_i=\lambda_ix_i,\quad i\in [1,s],
\end{align}
with
$$
\prod_{i=1}^s\lambda_i=1.
$$
Moreover, if all $x^i$, $i\in[1,s]$ are normalized, that is, $x^i_j\geq 0$, and
$$
\dsum_{j=1}^{n_i}x^i_j=1,\quad i\in [1,s],
$$
then $x^i$ are unique.
\end{prp}

When $X$ is a homogeneous sequence with $n_i=n_0$. Then its matrix form $M_X$ and its addition form $V_X$ satisfy
\begin{align}\label{4.1.6}
V_X=V_r(M_X),
\end{align}
where $V_r(A)$ is the row stacking form of a matrix $A$. \footnote{Assume $A=(a_{i,j})\in {\cal M}_{m\times n}$, then its row stacking form is
$$
V_r(A)=(a_{1,1},a_{1,2},\cdots a_{1,n}.\cdots, a_{m,1},\cdots,a_{m,n})^T\in \R^{mn}.
$$
}

\subsection{Normalized Matrices and Vectors}

\begin{dfn}\label{d4.2.1}
\begin{itemize}
\item[(i)] $x\in \R^k$ is called a stochastic vector if $x_i\geq 0$, $i\in [1,k]$ and
$$
\dsum_{i=1}^kx_i=1.
$$
The set of $k$-dimensional stochastic vectors is denoted by $\Upsilon^k$.
\item[(ii)] $X\in {\cal M}_{m\times n}$ is called a stochastic matrix if
$$
\Col_j(X)\in \Upsilon^m,\quad j\in [1,n].
$$
The set of $m\times n$-dimensional stochastic matrices is denoted by $\Upsilon_{m\times n}$.
\end{itemize}
\end{dfn}

\begin{exa}\label{e4.2.2}
\begin{itemize}
\item[(i)] Let $x\in \R^n$. Then
$$
softmax(x)\in \Upsilon^n
$$
is a stochastic vector.
\item[(ii)] Let $X\in {\cal M}_{s\times n_0}$. Then
$$
softmax(X)^T\in \Upsilon_{n_0\times s}.
$$
\end{itemize}
\end{exa}

The following result is obvious.

\begin{prp}\label{p4.2.3} If $A\in \Upsilon_{m\times n}$ and $x\in \Upsilon^{n}$, then
\begin{align}\label{4.2.1}
Ax\in \Upsilon^m.
\end{align}
\end{prp}

(\ref{4.2.1}) shows that for  a layer of MaxMul, if the transformation is a stochastic matrix and the input is a stochastic vector, then the output is also stochastic. No more softmax is necessary.

Define a weighted DK-STP as
\begin{dfn}\label{d4.2.4} \cite{che24} Let $A\in {\cal M}_{m\times n}$, $B\in {\cal M}_{p\times q}$, and $t=\lcm(n,p)$. Define
\begin{align}\label{4.2.2}
A\ttimes_w B:=\left(A\otimes \J^T_{t/n}\right) \left(B\otimes \frac{1}{t/p}\J_{t/p}\right).
\end{align}
\end{dfn}

Similarly to DK-STP, we have the following result \cite{che24}
\begin{prp}\label{p4.2.5}

\begin{align}\label{4.2.3}
A\ttimes_w B=A\Psi^w_{n\times p} B,
\end{align}
where
\begin{align}\label{4.2.4}
\Psi^w_{n\times p}=\frac{1}{t/p} \left(I_n\otimes \J^T_{t/n}\right)\left(I_p\otimes \J_{t/p}\right).
\end{align}
\end{prp}

Then it is easy to verify the following result.

\begin{prp}\label{p4.2.6}
\begin{itemize}
\item[(i)] Assume both $A$ and $B$ are stochastic matrices. Then $A\ttimes_w B$ is also a stochastic matrix.
\item[(ii)] Assume $A$ is a stochastic matrix and $x$ is a stochastic vector. Then $A\ttimes_w x$ is a stochastic vector.
\end{itemize}
\end{prp}

Proposition \ref{p4.2.6} is helpful for reducing the normalizing times.

\subsection{Operators over Hypervectors}

Consider $Q^{\mathrm{T}}=W^qX^{\mathrm{T}}$ (similarly, for $K$ and $V$). Then it is easy to verify the following result \cite{che12}

\begin{prp}\label{p4.3.1} Assume $X\in {\cal M}_{s\times n_0}$, then

\begin{align}\label{4.3.1}
V_{Q^{\mathrm{T}}}=\left(W^q\otimes I_{n_0}\right)V_{X^{\mathrm{T}}}.
\end{align}
\end{prp}

In the following we define some operators over hypervectors. First, we define the addition of hypervectors.

\begin{dfn}\label{d4.3.3}
\begin{itemize}
\item[(i)] Assume $X\in \R^{m_1\ltimes \cdots\ltimes m_s}$, $Y\in \R^{n_1\ltimes \cdots\ltimes n_s}$. Then
 \begin{align}\label{4.3.4}
 X+_rY=
 \begin{bmatrix}
(x^1+_r y^1)^T\\
(x^2+_r y^2)^T\\
\vdots\\
(x^s+_r y^r)^T
\end{bmatrix}\in {\cal M}_{s\times r}.
\end{align}
\item[(ii)]
 Assume $X\in \R^{m_1\ltimes \cdots\ltimes m_s}$, $Y\in \R^{n_1\ltimes \cdots\ltimes n_t}$, and $k=\lcm(s,t)$. Then

\begin{align}\label{4.3.5}
X+_rY=(X\otimes \J_{k/s})+_r (Y\otimes \J_{k/t})\in {\cal M}_{k\times r}.
\end{align}
\end{itemize}
\end{dfn}

Secondly, we define the inner-product of two hypermatrices.

\begin{dfn}\label{d4.3.3} Let $X\in \R^{m_1\ltimes \cdots\ltimes m_s}$, $Y\in \R^{n_1\ltimes \cdots\ltimes n_t}$. Then
\begin{itemize}
\item[(i)] The inner-product of $X$ and $Y$ is defined by
\begin{align}\label{4.3.6}
\begin{array}{l}
X\odot Y:=\\
\begin{bmatrix}
\langle x^1,y^1\rangle_{{\cal V}}&\langle x^1,y^2\rangle_{{\cal V}}&\cdots&\langle x^1,y^t\rangle_{{\cal V}}\\
\langle x^2,y^1\rangle_{{\cal V}}&\langle x^2,y^2\rangle_{{\cal V}}&\cdots&\langle x^2,y^t\rangle_{{\cal V}}\\
\vdots&~&~&~\\
\langle x^s,y^1\rangle_{{\cal V}}&\langle x^s,y^2\rangle_{{\cal V}}&\cdots&\langle x^s,y^t\rangle_{{\cal V}}\\
\end{bmatrix}\in {\cal M}_{s\times t}.
\end{array}
\end{align}
\item[(ii)] Assume $w_{i,j}=\lcm(m_i,n_j)$, $i\in [1,s]$, $j\in [1,t]$. Define a weighted matrix as
\begin{align}\label{4.3.7}
W:=\begin{bmatrix}
\sqrt{w_{1,1}}&\sqrt{w_{1,2}}&\cdots&\sqrt{w_{1,t}}\\
\sqrt{w_{2,1}}&\sqrt{w_{2,2}}&\cdots&\sqrt{w_{2,t}}\\
\vdots&~&~&~\\
\sqrt{w_{s,1}}&\sqrt{w_{s,2}}&\cdots&\sqrt{w_{s,t}}\\
\end{bmatrix}\in {\cal M}_{s\times t}.
\end{align}

The weighted
 inner-product of $X$ and $Y$ is defined by
\begin{align}\label{4.3.8}
X\odot_w Y:=(X\otimes Y)\circ W\in {\cal M}_{s\times t}.
\end{align}
\end{itemize}
\end{dfn}

Note that if $m_i=n_j=d$, $i\in [1,s]$, $j\in [1.t]$, it is easy to verify that
\begin{align}\label{4.3.9}
\begin{array}{l}
X\odot Y=\frac{XY^T}{d}\\
X\odot_w Y=\frac{XY^T}{\sqrt{d}}.
\end{array}
\end{align}

Thirdly, we define a linear mapping over hypervectors.

\begin{dfn}\label{d4.3.4}
Assume $X\in \R^{n_1\ltimes \cdots\ltimes n_s}$, $A\in {\cal M}_{s\times s}$.
The linear mapping $\pi_A:\R^{n_1\ltimes \cdots\ltimes n_s}\ra \R^{n_1\ltimes \cdots\ltimes n_s}$, denoted by $A\diamond X$ is defined as follows:
\begin{itemize}
\item Step 1: project-padding:
Replace $x^i$ by
\begin{align}\label{4.3.10}
\tilde{x}^i =\Pi^{n_i}_{n_0}x^i,\quad i\in [1,s],
\end{align}
where $n_0=\max\{n_i\;|\; i\in [1,s]\}$.

Then we have
$$
\tilde{X}=\begin{bmatrix}
(\tilde{x}^1)^T\\
(\tilde{x}^2)^T\\
\vdots\\
(\tilde{x}^s)^T\\
\end{bmatrix}\in {\cal M}_{s\times n_0}.
$$

\item Step 2:
\begin{align}\label{4.3.101}
\tilde{\Xi}:=A\tilde{X}.
\end{align}

\item Step 3:

Let
$$
\tilde{\xi}^i:=\left(\Row_i(\tilde{\Xi})\right)^T,\quad i\in [1,s].
$$
Define
\begin{align}\label{4.3.102}
{\xi}^i:=\Pi^{n_0}_{n_i}\tilde{\xi}^i,\quad i\in [1,s].
\end{align}
Then
\begin{align}\label{4.3.11}
A\diamond X:=\{{\xi}^1,{\xi}^2,\cdots, {\xi}^s\}\in \R^{n_1\ltimes \cdots\ltimes n_s}.
\end{align}
\end{itemize}
\end{dfn}

Since $\diamond$ is a linear mapping, we seek a matrix expression of the mapping $\pi_A$. It is convenient to express the related hypermatrices into their vector forms.

For nominal case,  assume $A\in {\cal M}_{s\times s}$ and $X\in {\cal M}_{s\times n_0}$.  Then it is easy verify the following result \cite{che12}

\begin{prp}\label{p4.3.5} Assume $X\in {\cal M}_{s\times n_0}$, then
$A\diamond X=AX$, and
\begin{align}\label{4.3.12}
V_{AX}=\left(A\otimes I_{s}\right)V_X.
\end{align}
\end{prp}

Now, assume $X\in \R^{n_1\ltimes \cdots\ltimes n_s}$. It is easy to verify that the Step 1 in Definition \ref{d4.3.4} can be expressed as
\begin{align}\label{4.3.13}
V_{\tilde{X}}=T_{\tilde{X}}V_X,
\end{align}
where
\begin{align}\label{4.3.14}
T_{\tilde{X}}=
\begin{bmatrix}
\Pi^{n_1}_{n_0}&0&\cdots&0\\
0&\Pi^{n_2}_{n_0}&\cdots&0\\
\vdots&~&~&~\\
0&0&\cdots&\Pi^{n_s}_{n_0}\\
\end{bmatrix}
\end{align}

Then the Step 2 in Definition \ref{d4.3.4} can be realized by
\begin{align}\label{4.3.15}
V_{\tilde{\Xi}} =(A\otimes I_s) V_{\tilde{X}}.
\end{align}

Finally, Step 3 in Definition \ref{d4.3.4} is performed by
\begin{align}\label{4.3.16}
V_{A\diamond  X} =T_{\Xi} V_{\tilde{\Xi}},
\end{align}
where
\begin{align}\label{4.3.17}
T_{\Xi} =\begin{bmatrix}
\Pi^{n_0}_{n_1}&0&\cdots&0\\
0&\Pi^{n_0}_{n_2}&\cdots&0\\
\vdots&~&~&~\\
0&0&\cdots&\Pi^{n_0}_{n_s}\\
\end{bmatrix}.
\end{align}

Summarizing the above three steps the whole process of the mapping $\pi_A$ can be described as follows.
\begin{align}\label{4.3.18}
V_X\xrightarrow{T_{\tilde{X}}} V_{\tilde{X}} \xrightarrow{A\otimes I_{s}} V_{\tilde{\Xi}} \xrightarrow{T_{\Xi}} V_{A\diamond X}.
\end{align}

Alternatively, it is formulated by
\begin{align}\label{4.3.19}
V_{A\diamond X}=T_{\Xi}(A\otimes I_s)T_{\tilde{X}}V_X.
\end{align}

\begin{exa}\label{e4.3.6} Assume
$$
X=\left[\begin{array}{ccc}
x^1_1&x^1_2&~\\
x^2_1&z^2_2&x^2_3\\
\end{array}\right);\quad
W=\begin{bmatrix}
w_{1,1}&w_{1,2}\\
w_{2,1}&w_{2,2}\\
\end{bmatrix}.
$$
We calculate $W\diamond X$ as follows.
\begin{itemize}
\item Step 1:
$$
V_X=(x^1_1,x^1_2,x^2_1,x^2_2,x^2_3)^T.
$$
$$
T_{\tilde{X}}=
\begin{bmatrix}
\Pi^2_3&0\\
0&I_3\\
\end{bmatrix}
=\begin{bmatrix}
1&0&0&0&0\\
0.5&0.5&0&0&0\\
0&1&0&0&0\\
0&0&1&0&0\\
0&0&0&1&0\\
0&0&0&0&1\\
\end{bmatrix}.
$$
$$
V_{\tilde{X}}=T_{\tilde{\Xi}}V_X
=(x^1_1,\frac{x^1_1+x^1_2}{2},x^1_2,x^2_1,x^2_2,x^2_3)^T.
$$

\item Step 2:
$$
\begin{array}{l}
W\otimes I_3\\
=\begin{bmatrix}
w_{1,1}&0&0&w_{1,2}&0&0\\
0&w_{1,1}&0&0&w_{1,2}&0\\
0&0&w_{1,1}&0&0&w_{1,2}\\
w_{2,1}&0&0&w_{2,2}&0&0\\
0&w_{2,1}&0&0&w_{2,2}&0\\
0&0&w_{2,1}&0&0&w_{2,2}\\
\end{bmatrix}
\end{array}
$$

$$
\begin{array}{l}
V_{\tilde{\Xi}}=(W\otimes I_3)V_{\tilde{X}}=\\
\left[w_{1,1}x^1_1+w_{1,2}x^2_1, \frac{1}{2}w_{1,1}(x^1_1+x^1_2)+w_{1,2}x^2_2,\right.\\
w_{1,1}x^1_2+w_{1,2}x^2_3, w_{2,1}x^1_1+w_{2,2}x^2_1,\\
\left.  \frac{1}{2}w_{2,1}(x^1_1+x^1_2)+w_{2,2}x^2_2,
w_{2,1}x^1_2+w_{2,2}x^2_3\right].
\end{array}
$$

\item Step 3:

$$
T_{\Xi}=
\begin{bmatrix}
\Pi^3_2&0\\
0&I_3\\
\end{bmatrix}
=\begin{bmatrix}
2/3&1/3&&0&0&0&0\\
0&1/3&2/3&0&0&0\\
0&0&0&1&0&0\\
0&0&0&0&1&0\\
0&0&0&0&0&1\\
\end{bmatrix}.
$$
$$
\begin{array}{l}
V_{W\diamond X}=T_{\Xi}V_{\tilde{\Xi}}=\\
\left[\frac{2}{3}(w_{1,1}x^1_1+w_{1,2}x^2_1)+\frac{1}{6}(w_{1,1}(x^1_1+x^1_2)+2w_{1,2}x^2_2),\right.\\
\frac{1}{6}(w_{1,1}(x^1_1+x^1_2)+2w_{1,2}x^2_2)
+\frac{2}{3}(w_{1,1}x^1_2+w_{1,2}x^2_3),\\
 w_{2,1}x^1_1+w_{2,2}x^2_1, \frac{1}{2}w_{2,1}(x^1_1+x^1_2)+w_{2,2}x^2_2,\\
\left.  w_{2,1}x^1_2+w_{2,2}x^2_3\right]^T.
\end{array}
$$

That is,
$$
\Xi:=W\diamond X=\left[\begin{array}{ccc}
q^1_1&q^1_2&~\\
q^2_1&q^2_2&q^2_3\\
\end{array}\right),
$$
where
$$
\begin{array}{l}
q^1_1=\frac{2}{3}(w_{1,1}x^1_1+w_{1,2}x^2_1)+\frac{1}{6}(w_{1,1}(x^1_1+x^1_2)+2w_{1,2}x^2_2),\\
q^1_2=\frac{1}{6}(w_{1,1}(x^1_1+x^1_2)+2w_{1,2}x^2_2)
+\frac{2}{3}(w_{1,1}x^1_2+w_{1,2}x^2_3),\\
 q^2_1=w_{2,1}x^1_1+w_{2,2}x^2_1\\
 q^2_2=\frac{1}{2}w_{2,1}(x^1_1+x^1_2)+w_{2,2}x^2_2,\\
q^2_3= w_{2,1}x^1_2+w_{2,2}x^2_3.
\end{array}
$$
\end{itemize}
\end{exa}

\section{Dimension-Free Transformer}

This section deals with hypervectors with varying dimensions. We will construct each building bricks for dimension-varying case.

\subsection{Input/Output Embeddion}

In general case, we have
\begin{align}\label{5.1.1}
\begin{array}{l}
X\in \R^{n_1\ltimes \cdots \ltimes n_{s}},\\
Q\in \R^{m_1\ltimes \cdots \ltimes m_{s}},\\
K\in \R^{a_1\ltimes \cdots \ltimes a_{s}},\\
V\in \R^{b_1\ltimes \cdots \ltimes b_{s}}.
\end{array}
\end{align}

\begin{itemize}
\item Nominal Case:
\end{itemize}

Assume
$$
n_{\a}=m_{\b}=a_{\gamma}=b_{\mu}:=d,\quad \forall \a,\b,\gamma,\mu.
$$
Then $X\in {\cal M}_{s_0\times d}$, and we have
\begin{align}\label{5.1.2}
\begin{array}{l}
Q^{\mathrm{T}}=W^qX^{\mathrm{T}},\quad W^q\in {\cal M}_{s\times d},\\
K^{\mathrm{T}}=W^kX^{\mathrm{T}},\quad W^k\in {\cal M}_{s\times d},\\
V^{\mathrm{T}}=W^vX^{\mathrm{T}},\quad W^v\in {\cal M}_{s\times d},\\
\end{array}
\end{align}
where $W^q,~W^k,~W^v\in {\cal M}_{s\times s}$.

Note that the values of $W^q$, $W^k$, and $W^v$ are time-varying through learning, but the size of these matrices  is fixed.

\begin{itemize}
\item General Case:
\end{itemize}

Next, we deal with general case where $x^{\a}$, $q^{\b}$, $k^{\gamma}$, and $v^{\mu}$ are all of different dimensions.
We consider $Q$ only. $K$ and $V$ can be treated in exactly the same way.

\vskip 2mm

First, we consider zero-padding, which is commonly used nowadays.
The following algorithm shows this.

\vskip 2mm

\begin{alg}\label{a5.1.1} Consider $X\in \R^{n_1\ltimes \cdots\ltimes n_{s_0}}$, and $Q\in \R^{m_1\ltimes \cdots\ltimes m_{s_q}}$.
\begin{itemize}
\item Step 1. Assume $d=\max\{n_{\a},m_{\b}\;|\;\a, \b \in [1,s]\}$.
Define the zero-padded $X$ as
\begin{align}\label{5.1.3}
\tilde{X}:=\{\tilde{x}_{i,j}\;|\;i\in[1,s],j\in[1,d]\},
\end{align}
where
$$
\tilde{x}_{i,j}=\begin{cases}
x_{i,j},\quad j\leq n_i,\\
0,\quad j>n_i.\\
\end{cases}
$$

\item Step 2.
Calculate padded $Q$, denoted by $\tilde{Q}$, as
\begin{align}\label{5.1.4}
\tilde{Q}^{\mathrm{T}}=W^q\tilde{X}^{\mathrm{T}}.
\end{align}

\item Step 3. Mask $\tilde{Q}$ to get $Q$ as
\begin{align}\label{5.1.5}
\begin{array}{l}
Q=
\begin{bmatrix}
(q^1)^T\\
(q^2)^T\\
\vdots\\
(q^{s_q})^T\\
\end{bmatrix}\\
\in \R^{m_1\ltimes \cdots\ltimes m_{s}},
\end{array}
\end{align}
where
$$
q^i_j=\tilde{q}_{i,j},\quad j\leq m_i,\quad i\in [1,s].
$$
\end{itemize}
\end{alg}

\vskip 2mm

Second, we propose a new technique, called the  project-padding, which is a building brick of DFT.
The following algorithm shows this.

\vskip 2mm

\begin{alg}\label{a5.1.2} Consider $X\in \R^{n_1\ltimes \cdots\ltimes n_{s}}$, and $Q\in \R^{m_1\ltimes \cdots\ltimes m_{s}}$.
\begin{itemize}
\item Step 1.  Assume the nominal dimension is $d$.
Define the project-padded $X$ as
\begin{align}\label{5.1.6}
\tilde{x}^i=\Pi^{n_i}_{d}x^i,\quad i\in [1,s],
\end{align}
and
\begin{align}\label{5.1.7}
\tilde{X}:=
\begin{bmatrix}
(\tilde{x}^1)^T\\
(\tilde{x}^2)^T\\
\vdots\\
(\tilde{x}^{s})^T\\
\end{bmatrix}
\end{align}

\item Step 2,
Calculate padded $\tilde{Q}$, as
\begin{align}\label{5.1.8}
\tilde{Q}=W^q\tilde{X}.
\end{align}

\item Step 3, Mask $\tilde{Q}$ to get $Q$ as
\begin{align}\label{5.1.9}
\begin{array}{l}
Q=
\begin{bmatrix}
(q^1)^T\\
(q^2)^T\\
\vdots\\
(q^{s_q})^T\\
\end{bmatrix}\\
\in \R^{m_1\ltimes \cdots\ltimes m_{s}},
\end{array}
\end{align}
where
$$
q^i=\Pi^d_{m_i}\tilde{q}^i,\quad i\in [1,s].
$$
\end{itemize}
\end{alg}

We use an example to describe these  padding/un-padding processes.

\begin{exa}\label{e5.1.3} Consider a sequence of batch size $s=4$ with $\dim(x^1)=\dim(x^4)=3$, $\dim(x^2)=4$, and $\dim(x^3)=5$. Assume the nominal dimension $n_0=6$, and
$$
W^q=(w_{i,j})\in {\cal M}_{n_0\times n_0}.
$$

Then we calculate $Q$:

\begin{itemize}
\item Zero-padding:
\end{itemize}

\begin{itemize}
\item Step 1:
$$
\tilde{X}=\begin{bmatrix}
x^1_1&x^1_2&x^1_3&0&0&0\\
x^2_1&x^2_2&x^2_3&x^2_4&0&0\\
x^3_1&x^3_2&x^3_3&x^3_4&x^3_5&0\\
x^4_1&x^4_2&x^4_3&0&0&0\\
\end{bmatrix}
$$
\item Step 2:
$$
\tilde{Q}^{\mathrm{T}}=W^q\tilde{X}^{\mathrm{T}}:=[\tilde{q}^1,\tilde{q}^2,\tilde{q}^3,\tilde{q}^4].
$$
where
$$
\tilde{q}^i=W^qx^i,\quad i\in[1,4].
$$

\item Step 3:
$$
Q=(q_{i,j})\in \R^{3\ltimes 4\ltimes 5\ltimes 3},
$$
where
$$
q_{i,j}=\tilde{q}_{i,j},\quad j\in [1,n_i], i\in[1,s].
$$
That is,
$$
\begin{array}{l}
q^1=\begin{bmatrix}
w_{1,1}x^1_1+w_{1,2}x^1_2+w_{1,3}x^1_3\\
w_{2,1}x^1_1+w_{2,2}x^1_2+w_{2,3}x^1_3\\
w_{3,1}x^1_1+w_{3,2}x^1_2+w_{3,3}x^1_3\\
\end{bmatrix},\\
q^2=\begin{bmatrix}
w_{1,1}x^2_1+w_{1,2}x^2_2+w_{1,3}x^2_3+w_{1,4}x^2_4\\
w_{2,1}x^2_1+w_{2,2}x^2_2+w_{2,3}x^2_3+w_{2,4}x^2_4\\
w_{3,1}x^2_1+w_{3,2}x^2_2+w_{3,3}x^2_3+w_{3,4}x^2_4\\
w_{4,1}x^2_1+w_{4,2}x^2_2+w_{4,3}x^2_3+w_{4,4}x^2_4\\
\end{bmatrix},\\
q^3=\begin{bmatrix}
w_{1,1}x^3_1+w_{1,2}x^3_2+w_{1,3}x^3_3+w_{1,4}x^3_4+w_{1,5}x^3_5\\
w_{2,1}x^3_1+w_{2,2}x^3_2+w_{2,3}x^3_3+w_{2,4}x^3_4+w_{2,5}x^3_5\\
w_{3,1}x^3_1+w_{3,2}x^3_2+w_{3,3}x^3_3+w_{3,4}x^3_4+w_{3,5}x^3_5\\
w_{4,1}x^3_1+w_{4,2}x^3_2+w_{4,3}x^3_3+w_{4,4}x^3_4+w_{4,5}x^3_5\\
w_{5,1}x^3_1+w_{5,2}x^3_2+w_{5,3}x^3_3+w_{5,4}x^3_4+w_{5,5}x^3_5\\
\end{bmatrix},\\
q^4=\begin{bmatrix}
w_{1,1}x^4_1+w_{1,2}x^4_2+w_{1,3}x^4_3\\
w_{2,1}x^4_1+w_{2,2}x^4_2+w_{2,3}x^4_3\\
w_{3,1}x^4_1+w_{3,2}x^4_2+w_{3,3}x^4_3\\
\end{bmatrix}\\
\end{array}
$$

\end{itemize}

\begin{itemize}
\item Projection-padding:
\end{itemize}
\begin{itemize}
\item Step 1:

Since
$$
\begin{array}{l}
\Pi^3_6=I_6(I_3\otimes \J_2)\\
~=\begin{bmatrix}
1&0&0\\
1&0&0\\
0&1&0\\
0&1&0\\
0&0&1\\
0&0&1\\
\end{bmatrix}.
\end{array}
$$
So
$$
\begin{array}{l}
\Pi^3_6(x^1)=(x^1_1,x^1_1,x^1_2,x^1_2,x^1_3,x^1_3),\\
\Pi^3_6(x^4)=(x^4_1,x^4_1,x^4_2,x^4_2,x^4_3,x^4_3).
\end{array}
$$
Similarly,
$$
\begin{array}{l}
\Pi^4_6=\frac{6}{12}(I_6\otimes \J^T_2)(I_4\otimes \J_3)\\
~=\frac{1}{2}\begin{bmatrix}
2&0&0&0\\
1&1&0&0\\
0&2&0&0\\
0&0&2&0\\
0&0&1&1\\
0&0&0&2\\
\end{bmatrix}.
\end{array}
$$
So
$$
\Pi^4_6(x^2)=(x^2_1,\frac{x^2_1+x^2_2}{2},x^2_2,x^2_3,\frac{x^2_3+x^2_4}{2},x^2_4).
$$
$$
\begin{array}{l}
\Pi^5_6=\frac{6}{30}(I_6\otimes \J^T_5)(I_5\otimes \J_6)\\
~=\frac{1}{5}\begin{bmatrix}
5&0&0&0&0\\
1&4&0&0&0\\
0&2&3&0&0\\
0&0&3&2&0\\
0&0&0&4&1\\
0&0&0&0&5\\
\end{bmatrix}.
\end{array}
$$
So
$$
\begin{array}{l}
\Pi^4_6(x^3)=\\
~(x^3_1,\frac{x^3_1+4x^3_2}{5},\frac{2x^3_2+3x^3_3}{5},\frac{3x^3_3+2x^3_4}{5},\frac{4x^3_4+x^3_5}{5},x^3_5).
\end{array}
$$
Finally, we have
$$
\begin{array}{l}
\tilde{X}=\\
\begin{bmatrix}
x^1_1&x^1_1&x^1_2&x^1_2&x^1_3&x^1_3\\
x^2_1&\frac{x^2_1+x^2_2}{2}&x^2_2&x^2_3&\frac{x^2_3+x^2_4}{2}&x^2_4\\
x^3_1&\frac{x^3_1+4x^3_2}{5}&\frac{2x^3_2+3x^3_3}{5}&\frac{3x^3_3+2x^3_4}{5}&\frac{4x^3_4+x^3_5}{5}&x^3_5\\
x^4_1&x^4_1&x^4_2&x^4_2&x^4_3&x^4_3\\
\end{bmatrix}
\end{array}
$$

\item Step 2:
$$
\tilde{Q}^{\mathrm{T}}=W^q\tilde{X}^{\mathrm{T}}:=[\tilde{q}^1,\tilde{q}^2,\tilde{q}^3,\tilde{q}^4]^{\mathrm{T}}\in {\cal M}_{4\times 6},
$$
where
$$
\begin{array}{l}
\tilde{q}^1_{i}=\dsum_{k=1}^6w_{i,k}\tilde{x}^1_k
=(w_{i,1}+w_{i,2})x^1_1\\
~+(w_{i,3}+w_{i,4})x^1_2+(w_{i,5}+w_{i,6})x^1_3,\\
\tilde{q}^2_{i}=\dsum_{k=1}^6w_{i,k}\tilde{x}^2_k
=(w_{i,1}+\frac{1}{2}w_{i,2})x^2_1+(\frac{1}{2}w_{i,2}\\
~+w_{i,3})x^2_2+(w_{i,4}+\frac{1}{2}w_{i,5})x^2_3+(\frac{1}{2}w_{i,5}+w_{i,6})x^2_4,\\
\tilde{q}^3_{i}=\dsum_{k=1}^6w_{i,k}\tilde{x}^3_k
=(w_{i,1}+\frac{1}{5}w_{i,2})x^3_1\\
~+(\frac{4}{5}w_{i,2}+\frac{2}{5}w_{i,3})x^3_2+(\frac{3}{5}w_{i,3}+\frac{3}{5}w_{i,4})x^3_3\\
~+(\frac{1}{5}w_{i,5}+w_{i,6})x^3_5,\\
\tilde{q}^4_{i}=\dsum_{k=1}^6w_{i,k}\tilde{x}^4_k=(w_{i,1}+w_{i,2})x^4_1\\
~+(w_{i,3}+w_{i,4})x^4_2+(w_{i,5}+w_{i,6})x^4_3.\\
\end{array}
$$

\item Step 3:
$$
Q=
\begin{bmatrix}
[(q^1)^T\\
(q^2)^T\\
(q^3)^T\\
(q^4)^T]^T
\end{bmatrix},
$$
Then
$$
q^i=\Pi^6_{n_i}\tilde{q}^i,\quad i\in [1,4].
$$
Now,
$$
\Pi^6_3=\frac{1}{2}
\begin{bmatrix}
1&1&0&0&0&0\\
0&0&1&1&0&0\\
0&0&0&0&1&1\\
\end{bmatrix},
$$
then
$$
q^1=\frac{1}{2}\begin{bmatrix}
\mu_{1,1}&\mu_{1,2}&\mu_{1,3}\\
\mu_{2,1}&\mu_{2,2}&\mu_{2,3}\\
\mu_{3,1}&\mu_{3,2}&\mu_{3,3}\\
\end{bmatrix}
\begin{bmatrix}
x^1_1\\
x^1_2\\
x^1_3
\end{bmatrix},
$$
$$
q^4=\frac{1}{2}\begin{bmatrix}
\mu_{1,1}&\mu_{1,2}&\mu_{1,3}\\
\mu_{2,1}&\mu_{2,2}&\mu_{2,3}\\
\mu_{3,1}&\mu_{3,2}&\mu_{3,3}\\
\end{bmatrix}
\begin{bmatrix}
x^4_1\\
x^4_2\\
x^4_3
\end{bmatrix},
$$
where
$$
\begin{array}{l}
\mu_{1,1}=w_{1,1}+w_{1,2}+w_{2,1}+w_{2,2}\\
\mu_{1,2}=w_{1,3}+w_{1,4}+w_{2,3}+w_{2,4}\\
\mu_{1,3}=w_{1,5}+w_{1,6}+w_{2,5}+w_{2,6}\\
\mu_{2,1}=w_{3,1}+w_{3,2}+w_{4,1}+w_{4,2}\\
\mu_{2,2}=w_{3,3}+w_{3,4}+w_{4,3}+w_{4,4}\\
\mu_{2,3}=w_{3,5}+w_{3,6}+w_{4,5}+w_{4,6}\\
\mu_{3,1}=w_{5,1}+w_{5,2}+w_{6,1}+w_{6,2}\\
\mu_{3,2}=w_{5,3}+w_{5,4}+w_{6,3}+w_{6,4}\\
\mu_{3,3}=w_{5,5}+w_{5,6}+w_{6,5}+w_{6,6}.\\
\end{array}
$$
Since
$$
\Pi^6_4=\frac{1}{3}
\begin{bmatrix}
2&1&0&0&0&0\\
0&1&2&0&0&0\\
0&0&0&2&1&0\\
0&0&0&0&1&2\\
\end{bmatrix},
$$
then
$$
q^2=\frac{1}{3}\begin{bmatrix}
\lambda_{1,1}&\lambda_{1,2}&\lambda_{1,3}&\lambda_{1,4}\\
\lambda_{2,1}&\lambda_{2,2}&\lambda_{2,3}&\lambda_{2,4}\\
\lambda_{3,1}&\lambda_{3,2}&\lambda_{3,3}&\lambda_{3,4}\\
\lambda_{4,1}&\lambda_{4,2}&\lambda_{4,3}&\lambda_{4,4}\\
\end{bmatrix}
\begin{bmatrix}
x^2_1\\
x^2_2\\
x^2_3\\
x^2_4
\end{bmatrix},
$$
where
$$
\begin{array}{l}
\lambda_{1,1}=2w_{1,1}+w_{1,2}+w_{2,1}+\frac{1}{2}w_{2,2}\\
\lambda_{1,2}=w_{1,2}+2w_{1,3}+\frac{1}{2}w_{2,2}+w_{2,3}\\
\lambda_{1,3}=2w_{1,4}+w_{1,5}+w_{2,4}+\frac{1}{2}w_{2,5}\\
\lambda_{1,4}=w_{1,5}+2w_{1,6}+\frac{1}{2}w_{2,5}+w_{2,6}\\
\lambda_{2,1}=w_{2,1}+\frac{1}{2}w_{2,2}+2w_{3,1}+w_{3,2}\\
\lambda_{2,2}=\frac{1}{2}w_{2,2}+w_{2,3}+w_{3,2}+2w_{3,3}\\
\lambda_{2,3}=w_{2,4}+\frac{1}{2}w_{2,5}+2w_{3,4}+w_{3,5}\\
\lambda_{2,4}=\frac{1}{2}w_{2,5}+w_{2,6}+w_{3,5}+2w_{3,6}\\
\lambda_{3,1}=2w_{4,1}+w_{4,2}+w_{5,1}+\frac{1}{2}w_{5,2}\\
\lambda_{3,2}=w_{4,2}+2w_{4,3}+\frac{1}{2}w_{5,2}+w_{5,3}\\
\lambda_{3,3}=2w_{4,4}+w_{4,5}+w_{5,4}+\frac{1}{2}w_{5,5}\\
\lambda_{3,4}=w_{4,5}+2w_{4,6}+\frac{1}{2}w_{5,5}+w_{5,6}\\
\lambda_{4,1}=w_{5,1}+\frac{1}{2}w_{5,2}+2w_{6,1}+w_{6,2}\\
\lambda_{4,2}=\frac{1}{2}w_{5,2}+w_{5,3}+w_{6,2}+2w_{6,3}\\
\lambda_{4,3}=w_{5,4}+\frac{1}{2}w_{5,5}+2w_{6,4}+w_{6,5}\\
\lambda_{4,4}=\frac{1}{2}w_{5,5}+w_{5,6}+w_{6,5}+2w_{6,6}\\
\end{array}
$$
Since
$$
\Pi^6_5=\frac{1}{6}
\begin{bmatrix}
5&1&0&0&0&0\\
0&4&2&0&0&0\\
0&0&3&3&0&0\\
0&0&0&2&4&0\\
0&0&0&0&1&5\\
\end{bmatrix},
$$
then
$$
q^3=\frac{1}{30}\begin{bmatrix}
\eta_{1,1}&\eta_{1,2}&\eta_{1,3}&\eta_{1,4}&\eta_{1,5}\\
\eta_{2,1}&\eta_{2,2}&\eta_{2,3}&\eta_{2,4}&\eta_{2,5}\\
\eta_{3,1}&\eta_{3,2}&\eta_{3,3}&\eta_{3,4}&\eta_{3,5}\\
\eta_{4,1}&\eta_{4,2}&\eta_{4,3}&\eta_{4,4}&\eta_{4,5}\\
\eta_{5,1}&\eta_{5,2}&\eta_{5,3}&\eta_{5,4}&\eta_{5,5}\\
\end{bmatrix}
\begin{bmatrix}
x^3_1\\
x^3_2\\
x^3_3\\
x^3_4\\
x^3_5
\end{bmatrix},
$$
where
$$
\begin{array}{l}
\eta_{1,1}=25w_{1,1}+5w_{1,2}+5w_{2,1}+w_{2,2}\\
\eta_{1,2}=20w_{1,2}+10w_{1,3}+4w_{2,2}+2w_{2,3}\\
\eta_{1,3}=15w_{1,3}+15w_{1,4}+3w_{2,3}+3w_{2,4}\\
\eta_{1,4}=10w_{1,4}+20w_{1,5}+2w_{2,4}+4w_{2,5}\\
\eta_{1,5}=5w_{1,5}+25w_{1,6}+w_{2,5}+5w_{2,3}\\
\eta_{2,1}=20w_{2,1}+4w_{2,2}+10w_{3,1}+2w_{3,2}\\
\eta_{2,2}=16w_{2,2}+8w_{2,3}+8w_{3,2}+4w_{3,3}\\
\eta_{2,3}=12w_{2,3}+12w_{2,4}+6w_{3,3}+6w_{3,4}\\
\eta_{2,4}=8w_{2,4}+16w_{2,5}+4w_{3,4}+8w_{3,5}\\
\eta_{2,5}=4w_{2,5}+20w_{2,6}+2w_{3,5}+10w_{3,6}\\
\eta_{3,1}=15w_{3,1}+3w_{3,2}+15w_{4,1}+3w_{4,2}\\
\eta_{3,2}=12w_{3,2}+6w_{3,3}+12w_{4,2}+6w_{4,3}\\
\eta_{3,3}=9w_{3,3}+9w_{3,4}+9w_{4,3}+9w_{4,4}\\
\eta_{3,4}=6w_{3,4}+12w_{3,5}+4w_{4,4}+12w_{4,5}\\
\eta_{3,5}=3w_{1,5}+15w_{5,6}+3w_{6,5}+3w_{6,6}\\
\eta_{4,1}=10w_{4,1}+2w_{4,2}+20w_{5,1}+4w_{5,2}\\
\eta_{4,2}=8w_{4,2}+4w_{4,3}+16w_{5,2}+8w_{5,3}\\
\eta_{4,3}=6w_{4,3}+6w_{4,4}+12w_{5,3}+12w_{5,4}\\
\eta_{4,4}=4w_{4,4}+8w_{4,5}+8w_{5,4}+16w_{5,5}\\
\eta_{4,5}=2w_{4,5}+10w_{4,6}+4w_{5,5}+20w_{5,6}\\
\eta_{5,1}=5w_{5,1}+w_{5,2}+25w_{6,1}+5w_{6,2}\\
\eta_{5,2}=4w_{5,2}+2w_{5,3}+20w_{6,2}+10w_{6,3}\\
\eta_{5,3}=3w_{5,3}+3w_{5,4}+15w_{6,3}+15w_{6,4}\\
\eta_{5,4}=2w_{5,4}+4w_{5,5}+10w_{6,4}+20w_{6,5}\\
\eta_{5,5}=w_{5,5}+5w_{5,6}+5w_{6,5}+25w_{6,6}\\
\end{array}
$$

\end{itemize}
\end{exa}

\begin{rem}\label{r5.1.4}
Comparing with zero-padding, projection-padding has the following advantages:
\begin{itemize}
\item[(i)] No additional junk information caused by padded zero. Hence, the Padding Mask, commonly used to eliminate the junk information can be omitted.

\item[(ii)] Even if the sequences which are longer than the nominated length, that is, $n_i>n_0$, projection-padding can still handle it by a standard projection and without losing information.

\item[(iii)] Finally, the output sequences (i.e., vectors) can  have the same sizes as the input sequences if we wish.
\end{itemize}
\end{rem}

\subsection{Dimension-Varying Attention}

This subsection reconsiders the scaled dot-product attention for dimension-varying case. Assume
\begin{align}\label{5.2.1}
\begin{array}{l}
Q\in \R^{m_1\ltimes \cdots \ltimes m_s},\\
K\in \R^{a_1\ltimes \cdots \ltimes a_s},\\
V\in \R^{b_1\ltimes \cdots \ltimes b_s}.\\
\end{array}
\end{align}

Recall (\ref{3.2.2}), since the $Q,~K,~V$ are only hypervectors, the matrix product is not applicable for them. We propose the following revision.
\begin{align}\label{5.2.2}
A=softmax(Q\odot_w K)\in {\cal M}_{s\times s}.
\end{align}

\begin{rem}\label{r5.2.1}
\begin{itemize}
\item[(i)] When $n_i=a_j=b_k:=n$, $\forall i,j,k\in [1,s]$, the $A$ in (\ref{5.2.2}) is exactly the same as the $A$ in (\ref{3.2.2}). Hence, (\ref{5.2.2}) is a generalization of (\ref{3.2.2}).
\item[(ii)] If we use
\begin{align}\label{5.2.3}
A=softmax(Q\odot K),
\end{align}
then even if in unified dimension case, we have
\begin{align}\label{5.2.4}
A=softmax\left(\frac{QK^T}{n}\right),
\end{align}
which is different from (\ref{3.2.2}) by a scaling. A conjecture is, (\ref{5.2.4}) might be better than (\ref{3.2.2}).
\end{itemize}
\end{rem}

Next, we consider (\ref{3.2.1}). Assume $V\in \R^{b_1\ltimes \cdots\ltimes b_s}$, (\ref{3.2.1}) is replaced by
\begin{align}\label{5.2.5}
Attension(Q,K,V)=A\diamond V.
\end{align}

\begin{rem}\label{r5.2.2} In most general case, we assume
$$
\begin{array}{l}
Q\in \R^{m_1\ltimes \cdots \ltimes m_{p}},\\
K\in \R^{a_1\ltimes \cdots \ltimes a_{q}},\\
V\in \R^{b_1\ltimes \cdots \ltimes b_{r}}.\\
\end{array}
$$
Then
$$
A:=Q\odot K~(\mbox{or}~ Q\odot_w K)\in {\cal M}_{p\times q}.
$$
Assume $t=\lcm(q,r)$, then we can define
\begin{align}\label{5.2.6}
Attension(Q,K,V):=(A\otimes \J^T_{t/q})\diamond (\J_{t/r}\otimes V).
\end{align}
It is worth noting that when (\ref{4.3.102}) is used to perform (\ref{5.2.6}), (\ref{4.3.102}) should be read as
\begin{align}\label{5.2.7}
{\xi}^i:=\Pi^{(t/d)n_0}_{n_i}\tilde{\xi}^i,\quad i\in [1,s].
\end{align}
\end{rem}

\subsection{Dimension-Varying Multi-Head Attention}

We need only to consider the Concat \& Linear part, which is depicted in Fig.\ref{Fig.3.7}.

Assume $V‘_i\in \R^{b^i_1\ltimes \cdots\ltimes b^i_s}$, $i\in [1,\ell]$, where $\ell$ is the head number, and
$V‘_0\in \R^{b^0_1\ltimes \cdots\ltimes b^0_s}$

Denote the component-wise expression of $V'_i$ by
$$
V'_i=\{v_i^1,\cdots,v_i^k\},\quad i\in [1,\ell],
$$
and
$$
V'_0=\{v_0^1,\cdots,v_0^k\}.
$$
 Then the Concat can be performed by setting
\begin{align}\label{5.3.1}
v_0^k=v_1^k+_{b^0_1}v_2^k +_{b^0_1}+\cdots +_{b^0_1}v_{\ell}^k,\quad k\in [1,s],
\end{align}
here the nominal addition is define by (\ref{2.2.7}).

Alternatively, expressing $V'_i$ in vector form as $V_{V'_i}$, then we define
\begin{align}\label{5.3.2}
V^i_c:=
\begin{bmatrix}
\Pi^{b^i_1}_{b^0_1}&0&\cdots&0\\
0&\Pi^{b^i_2}_{b^0_2}&\cdots&0\\
\vdots&~&~&~\\
0&0&\cdots&\Pi^{b^i_s}_{b^0_s}\\
\end{bmatrix}V_{V'_i}\in \R^{b^0},
\end{align}
where $b^0=\dsum_{k=1}^sb^0_k$.

Then we have
 \begin{align}\label{5.3.3}
V_{V'_c}=\dsum_{i=1}^{\ell} V^i_c.
\end{align}

Finally, the ``Linear" part can be realized by a set of linear mappings
 \begin{align}\label{5.3.4}
\pi_j:v^j_c\ra v^j_{d},\quad j\in [1.s],
\end{align}
where $v^j_c$ is the $j$-th component of $V'_c$, and $v^j_d$ is the $j$-th component of the output $V'$.

\begin{rem}\label{r5.3.1}
In fact (\ref{5.3.3}) can be replaced by a weighted sum. Let a set of weights $w_i\geq 0$, $i\in [1,n]$ are given, (say, $\dsum_{i=1}^nw_i=1$). Then  (\ref{5.3.3}) can be replaced by
 \begin{align}\label{5.3.5}
V_{V'_c}=\dsum_{i=1}^nw_iV^i_c.
\end{align}
\end{rem}

\subsection{Dimension-Free Add \& Norm}

Assume $X\in \R^{n_1\ltimes \cdots\ltimes n_s}$  and $F(X)\in \R^{m_1\ltimes \cdots\ltimes m_s}$. Then
 \begin{align}\label{5.4.1}
X+F(X):=Z\in \R^{m_1\ltimes \cdots\ltimes m_s},
\end{align}
where
$$
z^i=x^i+_{m_i}f^i=\Pi^{n_i}_{m_i}x^i+f^i,\quad i\in [1,s].
$$

\subsection{Dimension-Free Feed-Forward}

We need a generalized nominal addition for hypervectors.

\begin{dfn}\label{d5.5.1} Let $X\in \R^{n_1\ltimes \cdots \ltimes n_s}$, $Y\in \R^{m_1\ltimes \cdots \ltimes m_s}$, and ${\bf r}:=\{r_1,\cdots,r_s\}$, the nominal addition of $X$ and $Y$ is defined as
\begin{align}\label{5.5.1}
X+_{{\bf r}}Y:=\{z^1,\cdots,z^s\},
\end{align}
where
$$
z^i=x^i+_{r_i}y^i,\quad i\in[1,s].
$$
\end{dfn}

\begin{dfn}\label{d5.5.2} Assume $X\in \R^{n_1\ltimes \cdots \ltimes n_s}$, $B_1\in \R^{\a_1\ltimes \cdots \ltimes \a_s}$,
$B_2\in \R^{\b_1\ltimes \cdots \ltimes \b_s}$, $W_i\in {\cal M}_{s\times s}$, $i=1,2$. Insteed of (\ref{3.6.1}), the dimension-free feed forward can be described as
\begin{align}\label{5.5.2}
FFN(X)=W_2\diamond ReLU(W_1\diamond X+_{{\bf n}} B_1)+_{{\bf n}}B_2,
\end{align}
where ${\bf n}=\{n_1,\cdots,n_s\}$
\end{dfn}

\section{Assembled Attention}

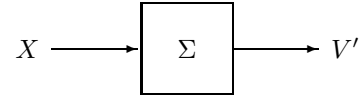
\begin{figure}
\centering
\setlength{\unitlength}{0.6 cm}
\begin{picture}(8,2)
\put(3,0){\framebox(2,2){$\Sigma$}}
\put(0.2,0.8){$X$}
\put(7.2,0.8){$V'$}
\put(1,1){\vector(1,0){1.9}}
\put(5,1){\vector(1,0){1.9}}
\end{picture}
\caption{Feed Forward\label{Fig.6.1}}
\end{figure}

To avoid complexity, we first assume the regular case, that is, $X,Q,K,V\in {\cal M}_{s\times n}$. Then a (self-) attention can be described into a black-box shown in Fig. \ref{Fig.6.1}.

In Fig. \ref{Fig.6.1} the attention is described as a mapping $\Sigma:{\cal M}_{s\times n}\ra {\cal M}_{s\times n}$, which is realized by
\begin{itemize}
\item Step 1:
$$
\begin{cases}
Q^{\mathrm{T}}=W^qX^{\mathrm{T}},\\
K^{\mathrm{T}}=W^kX^{\mathrm{T}},\\
V^{\mathrm{T}}=W^vX^{\mathrm{T}}.\\
\end{cases}
$$
\item Step 2:
$$
Attention(Q,K,V)=\frac{(QK^{\mathrm{T}})}{\sqrt{n}}V,\\
$$
\item Step 3:
$$
V'=softmax(Attention(Q,K,V)).
$$
\end{itemize}

The system parameters $W^q$, $W^k$, and $W^v$ can be determined through ``machine learning" using input-output date $\{(X(i),V'(i))\;|\; i=1,2,\cdots \}$. Now a natural question is: Can we use enough date to get unique set of parameters   $W^q$, $W^k$, and $W^v$? The answer is ``No!". Putting three steps together, we have
\begin{align}\label{6.1}
V'=softmax[X(W^q)^{\mathrm{T}}W^kX^{\mathrm{T}}X(W^v)^{\mathrm{T}}].
\end{align}
Now it is clear that
\begin{itemize}
\item[(i)] If
$$
W^{qk}:=(W^q)^{\mathrm{T}}W^k
$$
remains the same for different $W^q$ and $W^k$, the mapping $\Sigma$ remains the same.
\item[(ii)] If
$W^{qk}$ and $W^v$ are replaced simultaneously by
$\lambda W^{qk}$ and $\frac{1}{\lambda}W^v$, where $0\neq \lambda\in \R$,  the mapping $\Sigma$ remains the same.
\end{itemize}

\section{Conclusion}

Transformer is an attention flow based neural network. Nowadays, it is the key of some mostly attractive AI systems such as ChartGPT, DeepSeek, etc. Mathematically, it consists of different  matrix (including vector) products and additions with some nonlinear activation functions. The objects treated by AI are commonly of different dimensions. Then the padding and masking are used to overcome the trouble caused by dimension unmatching situation.

position Padding might cause some junk information, which is not completely deleted by masking.
STP and STA are dimension-free matrix/vector operators. They can be used naturally for dimension unmatching case without padding and masking. Using STP and STA, this paper has proposed  a dimension-free transformer. Each building bricks in a transformer have been reconstructed by STP/STA. Certain properties and algorithms have been developed. The STP/STA based AI system might be more efficient in various applications.

\centerline{ACKNOWLEDGMENT}

The author would like to thank Dr. Tiarong Zhang for fruitful discussions and suggestions which stimulated the paper.

\end{document}